\documentclass[twoside,11pt]{article}

\usepackage{blindtext}
\usepackage{makecell}
\usepackage{bm}

\usepackage[preprint]{jmlr2e}

\usepackage{microtype}
\usepackage{subcaption}
\usepackage{graphicx}
\usepackage{array}
\usepackage{multirow, makecell}
\usepackage{booktabs} 
\usepackage{xcolor}

\usepackage{amsmath}
\usepackage{amssymb}
\usepackage{mathtools}

\usepackage{algorithm}
\usepackage{algpseudocode}
\usepackage{xcolor}

\algrenewcommand\alglinenumber[1]{\textbf{#1}.}

\newcommand{\ud}{\,\mathrm{d}}

\newcommand{\R}{\mathbb{R}}

\newcommand{\cut}[1]{}
\newcommand{\der}[2]{\frac{\ud {#1}}{\ud {#2}}}

\DeclareMathOperator*{\argmin}{arg\,min}

\newcommand{\zt}{\tilde z}
\newcommand{\Zt}{\tilde Z}

\newtheorem{thrm}{Theorem}[section]

\usepackage{lastpage}
\jmlrheading{XX}{2026}{1-\pageref{LastPage}}{8/3; Revised X/XX}{X/XX}{21-0000}{Sudeepta Mondal, Xinyi Mary Xie, Alex Wong and Ganesh Sundaramoorthi}

\ShortHeadings{A Variational Information Theoretic Approach to Out-of-Distribution Detection}{Mondal, Xie, Wong \& Sundaramoorthi}
\firstpageno{1}

\begin{document}

\title{An Information-Theoretic Framework for Feature Construction in Out-of-Distribution Detection}

\author{\name Sudeepta Mondal \email        sudeepta.mondal2@rtx.com \\
       \addr RTX Technology Research Center\\
       East Hartford, CT 06118
       \AND
       \name Xinyi Mary Xie \email x.xie@yale.edu  \\
       \addr Department of Computer Science\\
       Yale University\\
       New Haven, CT 06511, USA
       \AND
       \name Alex Wong \email alex.wong@yale.edu\\
       \addr Department of Computer Science\\
       Yale University\\
       New Haven, CT 06511, USA
       \AND
       \name Ganesh Sundaramoorthi \email ganesh.sundaramoorthi@rtx.com \\
       \addr RTX Technology Research Center \\
       East Hartford, CT 06118}
       
\editor{My editor}

\maketitle

\begin{abstract}%
We present a theory for the construction of out-of-distribution (OOD) detection features for neural networks. We introduce random features for OOD through a novel information-theoretic loss functional consisting of two terms, the first based on the KL divergence separates resulting in-distribution (ID) and OOD feature distributions and the second term is the Information Bottleneck, which favors compressed features that retain the OOD information. We formulate a variational procedure to optimize the loss and obtain OOD features. Based on assumptions on OOD distributions, one can recover properties of existing OOD features, i.e., shaping functions. Furthermore, we show that our theory can predict a new shaping function that out-performs existing ones on OOD benchmarks, including both semantic and a wide range of covariate shifts. Our theory provides a general framework for constructing a variety of new features with clear explainability. 
\end{abstract}

\begin{keywords}
  Out-of-distribution detection, information-theoretic approaches, post-hoc methods, variational approaches
\end{keywords}

\section{Introduction}


Machine learning (ML) systems are typically designed under the assumption that the training and test sets are sampled from the same statistical distribution. However, this often does not hold in practice. For example, during deployment, test data may include previously unseen classes. In such cases, the ML system may produce incorrect results with high confidence \citep{devries2018learning}. Therefore, it is crucial to develop methods that enable ML systems to \emph{detect} \emph{out-of-distribution} (OOD) data. Detecting OOD data allows users to be alerted of potentially unreliable predictions and enables the system to adapt accordingly. OOD detection has gained considerable attention recently \citep{yang2022generalized}, and standardized benchmarks such as OpenOOD \citep{zhang2024openoodv15enhancedbenchmark} now enable systematic comparison across a large suite of detectors, networks, and OOD datasets. OOD inputs are commonly distinguished by the type of distributional shift: \emph{semantic} shift, where an input belongs to a previously unseen class, and \emph{covariate} shift, where an input retains an in-distribution class but differs in style, corruption, or domain, which can also lead to unpredictable results. It is important to detect all such distributional shifts \citep{yang2023full},\citep{yang2024imagenet}.

While there has been extensive work in the area of OOD detection~\citep{yang2022generalized,zhang2024openoodv15enhancedbenchmark}, recent state-of-the-art (SoA) \citep{sun2021react,djurisic2022ash,ahn2023line,sun2022dice, zhao2024towards, xu2023vravariationalrectifiedactivation, zhang2024learning} has focused on identifying descriptors of the data that can distinguish between OOD and ID data. Post-hoc approaches of OOD detection, which are applied to pre-trained models without additional training, have focused on constructing scoring functions to differentiate OOD from ID data. These detectors span several families, including score-based methods that derive a confidence signal from network outputs \citep{liu2021energy}, distance- or density-based methods \citep{sun2022knn}, and feature-shaping methods. In particular, feature-shaping functions have been shown to be promising. In feature shaping, features, usually from the penultimate layer of a pre-trained network, are input to a \emph{shaping} function and then used to score incoming data. Examples of these approaches include ReAct~\citep{sun2021react}, FS-OPT~\citep{zhao2024towards}, VRA~\citep{xu2023vravariationalrectifiedactivation}, ASH~\citep{djurisic2022ash}.  While these approaches have provided advancements to SoA on several benchmarks, most of these are empirically driven rule-based methods, and may not generalize well over new unseen datasets \citep{zhao2024towards}. Recent large-scale evaluations show that no single detector consistently outperforms others across OOD benchmarks \citep{zhang2024openoodv15enhancedbenchmark}. 
It is thus beneficial to understand under what conditions that these methods will work so that we may understand when to employ one method over another. One way to do this is to develop a theory where one can derive features for OOD detection (henceforth called OOD features for brevity) as a function of underlying assumptions (e.g., statistical distributions). This could potentially offer a way to critically examine existing methods. It could also potentially enable the systematic development of new features that may generalize better.

Towards this goal, we develop a theory to formulate OOD features based on underlying statistical distributions of ID and OOD distributions. We develop a novel loss functional, based on information theory, defined on the set of OOD features whose optimization yields OOD features as a function of the underlying statistical distributions. Unlike current approaches, our OOD features are random and thus follow a statistical distribution. The mean value models the deterministic shaping features in the literature.

Our loss aims to determine the OOD feature that maximally separates resulting ID and OOD feature distributions through the Kullback-Leiber (KL) divergence. As separating distributions by itself is ill-posed, we propose a novel use of the Information Bottleneck (IB) \citep{tishby2000information} as regularization. In our use, IB seeks compressed features that preserve the information the data has about OOD, aiming for a feature representation that contains only the information necessary for OOD detection. As this loss functional is defined on probability measures (representing the distribution of the OOD feature), it is an infinite dimensional optimization problem, and thus we use the calculus of variations \citep{troutman2012variational} to derive the optimization procedure. Our theory offers an explanation of several techniques employed in SoA rule-based approaches, and suggests a new shaping function that out-performs other shaping functions in SoA.

There have been recent theories for OOD detection \citep{zhao2024towards, xu2023vravariationalrectifiedactivation}. These works have introduced the novel idea of formulating OOD features through a loss function rather than empirically driven rule-based approaches of the past, and motivates our work.  In contrast to the aforementioned works, our theory employs a novel information-theoretic loss function, which offers several advantages. Our theory shows how different assumptions on the OOD distribution lead to different OOD feature shaping approaches.  Our theory is able to more accurately offer an explanation for properties of several SoA rule-based approaches as being from different underlying OOD distributions and different regularization (see next section for a more detailed discussion).

This paper is an extension of a preliminary version published as a conference version at ICML \citep{mondal2025variationalinformationtheoreticapproach}. In summary, our contributions are as follows: 
\begin{enumerate}
\item We introduce a novel theory and framework for deriving OOD features from neural networks. This involves the formulation of OOD features as a variational problem that formulates OOD features as \emph{random features} through a novel loss functional that contains two terms, one that maximizes the \emph{KL divergence} between the random feature under ID and OOD distributions and another term, the \emph{Information Bottleneck}, which extracts the information from the data that is relevant for OOD detection.
\item We develop the techniques to optimize the loss functional using the calculus of variations, and specifically derive a computationally feasible algorithm in the one-dimensional data case. 
\item Using our framework, we show how the OOD shaping functions change based on various data distributions.  We relate the mean value of our OOD features to existing OOD shaping functions. 
\item We introduce a novel piece-wise linear OOD feature shaping function (called PLF) predicted through our theory, and show that it leads to state-of-the-art results on OOD benchmarks. Beyond our conference paper, we formulate a new ``quantile'' formulation of PLF, which allows for a generalization of the assumption of stationary distributions across neural network features across channels without increasing the number of parameters of our approach.
\item We extend benchmarking beyond our conference paper to the OpenOOD benchmark \citep{zhang2024openoodv15enhancedbenchmark} and the BROAD dataset \citep{guille2024expecting}, which includes semantic shifts and covariate shifts, and compare against the post-hoc inference methods in OpenOOD. We demonstrate state of the art results in terms of generalization across datasets in these benchmarks - \textbf{out of 21 methods, our method ranks in the top three on all datasets/architectures, which no other method is able to achieve.} In particular, we show our method extends beyond sematic shifts to covariate shifts. We also perform the first comprehensive benchmarking of post-hoc OOD detectors in OpenOOD on covariate shifts.


\end{enumerate}

\subsection{Related Work}


We briefly review related work; the reader is referred to \cite{yang2022generalized} for a survey. Post-hoc OOD detectors have focused on constructing scoring functions to differentiate OOD from ID data, leveraging confidence scores \citep{hendrycks2018,zhang_decoupling,odin_liang}, energy-based metrics \citep{liu2021energy,wu2023energy,elflein2021out} and distance-based measures \citep{lee2018simple,sun2022knn}. For example, MSP \citep{hendrycks2018} used the maximum softmax probability as a confidence score. ODIN \citep{odin_liang} improved OOD detection by applying temperature scaling and adding small perturbations to input data before computing the maximum softmax probability. \citep{ren2019likelihood} proposes to use the likelihood ratio, which has been proposed over likelihoods, which do not work well \citep{kirichenko2020normalizing}. \citep{lee2018simple} leveraged Mahalnobis distance to compute the distance between features and classes. KNN \citep{sun2022knn} uses a non-parametric approach. Energy methods \citep{liu2021energy} present an alternative to softmax scores by employing the Helmholtz free energy. Energy scoring has been adopted by several \emph{OOD feature-shaping} approaches; feature-shaping is the focus of our work. \cite{magesh2023principled} show how to combine multiple scores in a principled manner.

{\bf Feature-shaping approaches to OOD detection}: Several methods perform OOD detection by computing features of the output of layers of the neural network \citep{sun2021react,Kong_BFAct,djurisic2022ash,fort2021exploring,zhao2024towards} before being input to a score. In ReAct \citep{sun2021react}, the penultimate layer outputs are processed element-wise by clipping large values. It is empirically noted that OOD data results in large spikes in activations, which are clipped to better separate the ID and OOD distributions. BFAct \citep{Kong_BFAct} uses the Butterworth filter to smoothly approximate the clipping. ASH computes features by sparsifying intermediate outputs of the network by flooring small values to zero and passing larger values with possible scaling. DICE \citep{sun2022dice} is another approach to sparsification. Different than element-wise approaches, ASH then does vector processing of the shaped feature before input to a score. VRA \citep{xu2023vravariationalrectifiedactivation} and \cite{zhanglearning} derive element-wise shaping functions by an optimization approach. 

{\bf Optimization-based approaches for feature shaping}: \cite{xu2023vravariationalrectifiedactivation} formulates a loss function for deterministic OOD features that aims to separate the means of ID and OOD feature distributions, and regularization is added to keep the OOD feature near the identity through the $L_2$ norm. \cite{zhao2024towards} analyzes a similar loss function but with point-wise rather than $L_2$ regularization. They further offer simplifications to remove the reliance on the OOD distribution. These works have introduced the novel idea of formulating OOD features through a loss function. Our approach offers several advantages.
Over \cite{zhao2024towards}, we present a framework in which we can study the OOD feature as a function of the underlying OOD distribution. This shows the implicit assumptions in several existing methods. In contrast, \cite{zhao2024towards} aims to remove dependence on the OOD distribution. Our results show that feature shaping can vary as a function of the underlying OOD distribution. Over \cite{zhao2024towards,xu2023vravariationalrectifiedactivation}, our theory offers an explanation of qualitative properties of existing SoA methods. For instance, clipping of large values in OOD features (of ReAct \citep{sun2021react}) is associated with a higher Information Bottleneck (IB) regularization which is needed for noisier OOD datasets. Negative slope at large values in \cite{zhao2024towards,xu2023vravariationalrectifiedactivation} is associated with low IB regularization. Also, pruning of small feature values in \cite{xu2023vravariationalrectifiedactivation,djurisic2022ash} is associated with OOD distributions with heavier tails. See Section~\ref{sec:ood_vs_distribution} for more technical details.

\section{Variational Formulation of OOD Features}


We formulate OOD features as an optimization problem. For the sake of the derivation, we will assume that the probability distributions of ID and OOD features from the network are given in this section. In practice, the ID distribution can be estimated by training data, and can thus be assumed available. In Section~\ref{sec:ood_vs_distribution}, we will then study the OOD features under various OOD distributional assumptions to show how features vary with distribution and offer plausible assumptions made by existing feature shaping approaches. We will also make reasonable assumptions on the OOD distribution to derive new prescriptive OOD features for use in practice.

Current OOD features in the OOD literature are computed by processing features from the neural network through a deterministic function (e.g., clipping). In contrast, we propose to generalize that approach by allowing for \emph{random} functions. This generalization naturally fits in the framework of probabilistic and information theoretic tools, and these tools are naturally suited for the probabilistic distributional assumptions we wish to study. Let $Z$ denote the feature (a random variable) from the network (penultimate or intermediate layer feature). We denote by $\Zt$ the random OOD feature (a random variable) that we seek to determine. The distribution of $\Zt$ is determined through the conditional distribution $p(\zt|z)$. Thus, rather than solving for a deterministic function $f(Z)$, we instead solve for a random feature $\Zt$ represented through $p(\zt|z)$ as in Information Theory \citep{cover1999elements}. Thus, given a feature $z$ sampled from the distribution of $Z$, the OOD feature is $\Zt \sim p(\zt|z)$. We will primarily be concerned with the mean value of the distribution in this paper to relate to existing feature shaping methods. Let $X$ be the random variable indicating the data (e.g., image, text), and $Y$ be the random variable indicating in- ($Y=0$) and out-of- ($Y=1$) distribution data. Note this forms a Markov Chain $Y \rightarrow X \rightarrow Z \rightarrow \Zt$. The Markov Chain property is needed to construct one of the terms of our loss function, discussed next.


We propose a novel loss functional to design the OOD random feature. This loss functional is defined on the space of conditional distributions $p(\zt|z)$. The first term aims to separate the ID and OOD distributions of the random feature $\Zt$. This is natural since we would like to use the OOD feature to better separate the data into in or out-of-distribution.  To achieve this separation, we propose to maximize the symmetrized KL-divergence between $p(\zt|Y=0)$ and $p(\zt|Y=1)$. Note \citep{zhao2024towards} also seeks to separate OOD feature distributions, however, differently than our approach as only the means of the distribution are separated. Also, note that $p(\zt|Y=y)$ is a function of $p(\zt|z)$, the variable of optimization, and thus the KL term is a function of $p(\zt|z)$. This term is defined as follows:
\begin{multline} \label{eq:KLterm}
    D_{KL}(p(\zt|z)) =
    D_{KL}[ p(\tilde z|Y=1)\, ||\, p(\tilde z | Y=0) ] + 
    D_{KL}[ p(\tilde z|Y=0)\, ||\, p(\tilde z | Y=1) ],
\end{multline}
where 
\begin{align}
    D_{KL}[ p \, ||\, q ] &= \int p(x)\log{\frac{p(x)}{q(x)}} \ud x,
    \quad \textrm{and} \\
    \quad p(\tilde z | y ) &= \int p(\tilde z | z) p(z|y) \ud z.
\end{align}
Note that we have used that $p(\tilde z|z,y)=p(\tilde z|z)$ as the OOD feature is constructed the same for both ID and OOD data. These equations shows the dependence of the OOD feature distributions on $p(\zt|z)$. The KL divergence is a natural choice for separating distributions and a standard information-theoretic quantity.

Unconstrained maximization of KL divergence is ill-posed since one may choose features that can be separated arbitrarily far, and thus the problem must be constrained. Also, it is possible to reduce the dimensions of $Z$ to a few dimensions whose distributions are separated arbitrarily far but remove information necessary to fully characterize OOD data. Therefore, we need to ensure that $\Zt$ contains all (and only) the information relevant to accurately determine OOD data. 

\begin{figure}
    \centering
    \includegraphics[width=1\linewidth]{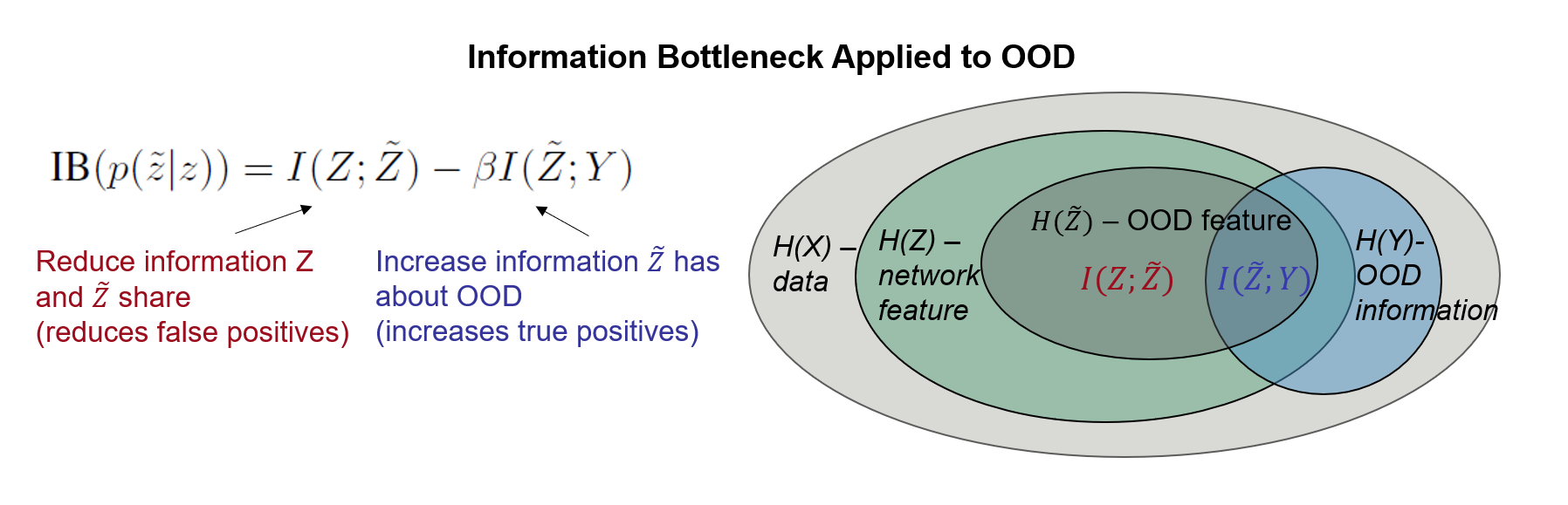}
    \caption{A conceptual diagram illustrating various information-theoretic quantities in the information bottleneck for the OOD feature computation problem. $H$ indicates the entropy of a random variable, and $I$ denotes the mutual information. The information pertaining to OOD is the entropy of $Y$. The network feature $Z$ reduces the information in the original data $X$ due to the data processing inequality.  The goal in designing a feature $\tilde Z$ to minimize the information bottleneck is to design a feature that captures only and all the information from $Y$.  This is done by decreasing the mutual information between $Z$ and $\tilde Z$ so that information that leads to false positives is reduced, while increasing the mutual information between $\tilde Z$ and $Y$ to increase true positives.}
    \label{fig:IB_schematic}
\end{figure}

With these considerations, we aim to compress the dimensions of $Z$ to form a simple/compact feature, but in a way that preserves the OOD information (contained in the variable $Y$). To achieve this, we adapt the Information Bottleneck \citep{tishby2000information}. In the Information Bottleneck method, the quantization of a random variable $X$ is considered to form the random variable $T$ in such a way to preserve information about a random variable $Y$, where $Y$ forms a Markov Chain with $X$, i.e., $Y\rightarrow T \rightarrow X$. A functional is formulated on all possible random variables such that when it is minimized, the optimized argument is the desired $T$. This is precisely the functional we would like to determine our OOD feature $\Zt$ (where $\Zt$ is analogous to the quantized variable $T$ and $Z$ is analogous to the original, non-quantized, random variable $X$). Thus, the second term of our functional, following from \citep{tishby2000information}, is 
\begin{equation} \label{eq:IB}
    \mbox{IB}(p(\tilde z|z)) = I(Z; \tilde Z) - \beta I(\tilde Z; Y),
\end{equation}
where $I$ indicates mutual information, and $\beta>0$ is a hyperparameter. The first term of \eqref{eq:IB} is the compression term that measures the mutual information between $Z$ and $\Zt$; this term is minimized and thus the term favors $\Zt$ to be a compressed version of $Z$. The second term maximizes the mutual information between $\Zt$ and $Y$, and thus favors $\Zt$ to retain OOD relevant information. A visualization of the quantities in the information bottleneck in the context of the OOD detection problem is shown in Figure~\ref{fig:IB_schematic}. It shows how the two mutual information terms favor constructing a feature $\tilde Z$ that aims to reduce false positives in OOD detection (arising from extraneous changes in the data that do not consist of changes that relate to OOD) and increase the true positive detections by ensuring the feature $\tilde Z$ increases information about OOD. Note the diagram also suggests why network features, rather than the the data itself is used for OOD detection in the literature, which is to reduce the variability in the data that correspond to changes that are not OOD (in the light grey area that do not overlap with the blue region), and hence reduce false positives. The OOD feature further reduces false positives by reducing the changes of the network feature that do not constitute OOD (overlapping with the blue region).

Thus, our combined loss functional is
\begin{equation} \label{eq:OODfunctional}
    L(p(\tilde z|z)) = -D_{KL}(p(\zt|z)) + \alpha \mbox{IB}(p(\tilde z|z)), 
\end{equation}
which is minimized to determine the conditional distribution of $\Zt$, $p(\zt|z)$, and $\alpha>0$ is a hyperparameter. Our goal is to determine the optimal $p(\zt|z)$, which can then be used with a score function to determine whether data $z$ is OOD or not. Note that we are seeking to optimize over the set of continuous probability distributions, which forms an infinite dimensional optimization problem. To gain intuition into the loss functional above, in particular to see that it forms a well-posed problem and that IB regularization is needed, we analyze a simple case with 1D Gaussian distributions that result in closed form solution in Appendix~\ref{app:1DGaussianLoss}.
We verify in the next section that the loss functional, for more complex distributions/features, yields well-posed problems and hence results in an optimal solution.

\section{Optimization for OOD Features}
\label{sec:optimization}
In this section, we discuss the optimization of the loss functional \eqref{eq:OODfunctional}. The loss functional is defined on continuous probability density functions $p(\zt|z)$, where $z, \zt$ are continuous. This is an infinite dimensional optimization problem, and to find the optimal feature one can use the calculus of variations to determine the gradient of $L$ \citep{troutman2012variational}. Setting the gradient to zero and solving for the probability distribution that satisfies the equation gives the necessary conditions for the optimizer. For our loss, that approach does not yield a closed form solution and so we instead use the gradient to perform a gradient descent.

\subsection{Loss Under Element-wise Independence of Feature}
Because formulating numerical optimization for general multi-dimensional distributions is difficult, we make some simplifications to gain insights to our theory and approach. Even with these simplifications, we will show that our approach can explain popular approaches in the literature and lead to a new state of the art method. Our first simplification (which is similar to element-wise processing assumptions made in existing methods, e.g., \citep{sun2021react,zhao2024towards}) is to assume that the conditional feature distribution $p(\zt|z)$ can be factorized as $p(\zt|z) = \prod_{i=1}^n p(\zt_i|z)$, which assumes conditional independence of the components of $\tilde z$ and that each component has the same conditional distribution. We also assume that $p(z|y)=\prod_{i=1}^n p(z_i|y)$, that is, the components of $z$ are independent conditioned on $y$. Under these assumptions, the optimization of the loss functional \eqref{eq:OODfunctional} reduces to the optimization of several optimization problems defined on one-dimensional probability distributions from each feature component (see Appendix~\ref{app:simplification_loss} for details):
\begin{align}
    \argmin_{p(\zt_i|z_i)} L_i(p(\zt_i|z_i)), \quad
    i\in \{1,\ldots, n\},
\end{align}
where
\begin{multline} \label{eq:1Dloss}
    L_i(p(\zt_i|z_i)) = 
    -D_{KL}[p(\zt_i|0)\,||\,p(\zt_i|1)]- 
    D_{KL}[p(\zt_i|1)\,||\,p(\zt_i|0)] + 
    \alpha [ I(\Zt_i; Z_i) -\beta I(\Zt_i;Y) ].
\end{multline}
Thus, we next provide an optimization procedure for each of the loss functionals above, defined on one-dimensional distributions. For simplicity of notation, we now omit the $i$ subscripts.

\subsection{Gradient of Loss Functional}
We will use gradient descent to optimize the loss functional. Since the problem is non-convex, gradient descent is a natural choice. Given the infinite dimensional problem, we use the calculus of variations to compute the gradient.

We perform the computation for the gradient of \eqref{eq:OODfunctional} in Appendix~\ref{app:gradients} and summarize the result in the following theorem:
\begin{thrm}[Gradient of Loss]
    The gradient of $D_{KL}(p(\zt|z)))$ \eqref{eq:KLterm} with respect to $p(\zt|z)$ is given (up to an additive function of $z$) by
    \begin{align} \label{eq:KL_grad}
         \nabla_{p(\tilde z|z)} D_{KL} &= 
    p(z|0) \cdot 
    \left[
        l(z)\log l(\tilde z) - l(\tilde z)
    \right] 
    - 
     p(z|1) \cdot 
    \left[
        l(z)^{-1}\log l(\tilde z) + l(\tilde z)^{-1}
    \right], 
    \end{align}
    where $p(z|y)=p(z|Y=y)$, $p(\zt|y)=p(\zt|Y=y)$ and  
    \begin{equation}
        l(z) = \frac{p(z|1)}{p(z|0)}, \quad 
        \textrm{and} \quad
        l(\zt) = \frac{p(\zt|1)}{p(\zt|0)}.
    \end{equation}
    The gradient of $\mbox{IB}(p(\zt|z))$ \eqref{eq:IB} is given by 
    \begin{equation} \label{eq:IB_grad}
    \nabla_{p(\tilde z|z)} \mbox{IB}=
    \sum_{y\in\{0,1\}} p(y) p(z|y) 
    \left[
    \log{\frac{p(\tilde z|z)}{p(\tilde z)}} - 
    \beta  \log{ \frac{p(\tilde z|y)}{p(\tilde z)} }
    \right].
    \end{equation}
    The gradient of the full loss $L$ in \eqref{eq:OODfunctional} is then
    \begin{equation} \label{eq:loss_grad}
        \nabla_{p(\zt|z)}L = -\nabla_{p(\zt|z)} D_{KL}+ \alpha \nabla_{p(\zt|z)} \mbox{IB}.
    \end{equation}
\end{thrm}

To simplify further and study a model that more closely resembles OOD feature shaping functions in the literature, we make the following assumption:
\begin{equation} \label{eq:1DGaussian_random_feature}
    p(\zt|z) \sim \mathcal{N}( \mu(z), \sigma_c(z) ), 
\end{equation}
where $\mathcal N$ indicates Gaussian distribution, $\zt,z\in \R$ and $\mu, \sigma_c : \R\to\R$ are the mean/standard deviation. We use the sub-script c to denote ``conditional" to distinguish it from other sigmas used below. We can think of this model as random perturbations of a deterministic feature shaping function $\mu$, in particular,
\begin{equation}
    \Zt = \mu(Z) + \varepsilon(Z), \quad \mbox{where } \varepsilon \sim \mathcal N(0,\sigma_c(Z)).
\end{equation}
The OOD's feature mean value for a given network feature $z$ is $\mu(z)$.  The closer $\sigma_c$ is to zero, the closer the approach is to deterministic feature shaping.   Note if the optimization turns out to result in $\sigma_c = 0$, then deterministic functions would be optimal. In our numerous simulations, this does not happen and thus random OOD features appear to be more optimal with respect to the loss that we defined. We now compute the gradients with respect to $\mu$ and $\sigma_c$: 
\begin{thrm}[Loss Gradient Under Gaussian Random OOD Feature \eqref{eq:1DGaussian_random_feature}]
    The gradient of the loss \eqref{eq:OODfunctional} under \eqref{eq:1DGaussian_random_feature} is
    \begin{align}
        \nabla_{\mu} L(z) &=
        \int
        \frac{\nabla_{p(\tilde z|z)}L(\zt,z)}{\sigma_c^2(z) } 
        [\tilde z-\mu(z)] p(\tilde z|z) 
        \ud \tilde z \\
        \nabla_{\sigma_c} L(z) &= 
        \int
        \frac{\nabla_{p(\zt|z)} L(\zt,z)}{\sigma_c(z)}
        \left[
        \frac{(\zt-\mu(z))^2}{\sigma_c^2(z)} - 1
        \right] p(\zt|z)  
        \ud \zt,
    \end{align}
    where $\nabla_{p(\zt|z)}L$ is given in \eqref{eq:loss_grad}.
    
\end{thrm}

\subsection{Numerical Optimization of Loss}
\label{subsec:numerical_optimization}
We implement a gradient descent algorithm using a discretization of the continuum equations above. We choose a uniform discretization of the space of $z$, i.e., $\{z_i\}_i\subset \R$.  We represent $\mu$ and $\sigma_c$ through their samples: $\mu_i=\mu(z_i)$ and $\sigma_{c,i} = \sigma_c(z_i)$. We specify formulas for $p(\zt)$ and $p(\zt|y)$ under the discretization, which will be required in the computation of the approximation to the gradient:
\begin{align}
    p(\tilde z|y) &= \sum_i p(\tilde z|z_i)p(z_i|y)\Delta z_i  
    = \sum_i \frac{1}{\sigma_{c,i}} G_{\sigma_{c,i}}(\tilde z-\mu_i) p(z_i|y) \Delta z_i\\
    p(\tilde z) &= \sum_y p(y) p(\tilde z|y).
\end{align}
Thus, $p(\zt|y)$ is approximated as a mixture of Gaussians. The gradient descent is shown in Algorithm~\ref{alg:numerical_gradient_descent}, which assumes ID and OOD distributions and determines the Gaussian random feature parameterized through $\mu$ and $\sigma_c$.  

The complexity for this optimization (which is done off-line in training) is $\mathcal{O}(NMK)$ where $N$ is the number of samples of $p(z|y)$, $M$ is the samples of $p(\tilde z|z)$ and $K$ is the number of gradient descent iterations.  On a single A100 GPU, this took less than a minute.

\begin{algorithm}[t]
\caption{1D Gaussian Random Feature Computation}
\label{alg:numerical_gradient_descent}
\begin{algorithmic}[1]\small
    \State \textbf{Input:} ID/OOD Distributions $p(z|y)$, hyper-paramters for ID, $\alpha, \beta$ and learning rate $\eta$
    \State \textbf{Output:} Optimal mean $\mu_i$ and standard deviation $\sigma_{c,i}$ for each $i$
    \State Initialize: $\mu_i = z_i$, and $\sigma_{c,i} = \text{const}$
    \For{$n$ iterations}
    \For{$i$}
    \State Compute a discretization of $\zt$ in its likely range: $\zt_i^j \in (\mu_i-k\sigma_{c,i}, \mu_i+k\sigma_{c,i})$ where $k\geq 3$
    \For{$j$}
    \State Compute the loss gradient with respect to $p(\zt|z)$:
                \begin{align*}
                \nabla_{p(\tilde z|z)} L(\tilde z^i_j, z_i) &=
                p(z_i|0) \cdot \left[
                    l(z_i)\log l(\tilde z^i_j) - l(\tilde z^i_j)
                \right] -
                p(z_i|1)  \cdot \left[
                    l(z_i)^{-1}\log l(\tilde z^i_j) + l(\tilde z^i_j)^{-1}
                \right] \\
                &+ \alpha 
                \sum_{y\in\{0,1\}}  p(y) p(z_i|y) \left[
                \log{\frac{p(\tilde z^i_j|z_i)}{p(\tilde z^i_j)}} - 
                \beta  \log{ \frac{p(\tilde z^i_j|y)}{p(\tilde z^i_j)} } 
                \right]
                \end{align*}
    \EndFor
    \State Compute the gradients of the loss with respect to $\mu,\sigma_c$:
                \begin{align*}
                 \nabla_{\mu} L(z_i) &=
                     \sum_j 
                     \frac{\nabla_{p(\tilde z|z)} L(\tilde z^i_j, z_i)}{\sigma_{c,i}^2}
                    (\tilde z^i_j-\mu_i)
                     p(\tilde z^i_j|z_i) \Delta z_i \\
                \nabla_{\sigma_c} L(z_i) &=
                \sum_j 
                    \frac{\nabla_{p(\tilde z|z)} L(\tilde z^i_j, z_i)}{\sigma_{c,i}}
                    \left[
                    \frac{(\zt_j^i-\mu_i)^2}{\sigma_{c,i}^2} - 1
                    \right] 
                    p(\tilde z^i_j|z_i) \Delta z_i
                \end{align*}
    \EndFor
    \For{$i$}
    \State Update $\mu_i, \sigma_{c,i}$:
    \begin{align*}
                    \mu_i &\leftarrow \mu_i - \eta
                    \nabla_{\mu} L(z_i)
                    \\
                    \sigma_{c,i} &\leftarrow 
                    \sigma_{c,i} - \eta
                    \nabla_{\sigma_c} L(z_i)
      \end{align*}
    \EndFor
    \EndFor
\end{algorithmic}

\end{algorithm}


\section{A Study of OOD Features vs Distribution}

\label{sec:ood_vs_distribution}

In this section, we study the resulting OOD features based on various choices of distributions using the algorithm in the previous section, and relate these choices to OOD feature shaping techniques that are present in the literature. Note that while in practice the OOD distribution is unknown, our theory nevertheless suggests the underlying distributional assumptions of prominent existing feature shaping methods. This is useful to understand when these methods will generalize as a function of the type of OOD data. We will also derive a generic OOD shaping function, encompassing properties of several distributions, and show that this shaping function can lead to SoA performance in the next section. Note in practice, we have observed that distributions from OOD datasets to exhibit similarities to the distributions studied, see Appendix~\ref{app:feature_distributions}. We will provide further rationale on studying these distributions below.

\begin{figure}
    \centering
    \includegraphics[width=0.9\linewidth]{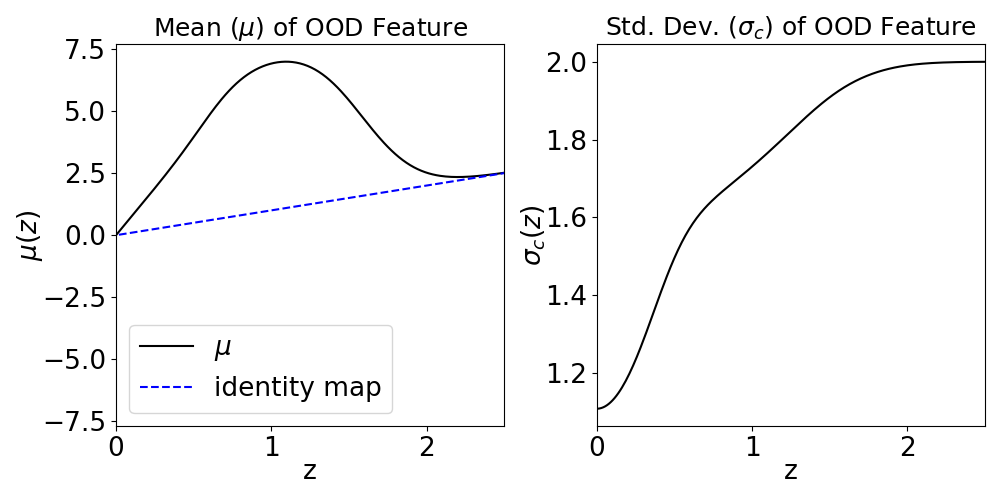}
    \caption{Optimal OOD Gaussian Feature Under Example Gaussian ID/OOD Distributions. The optimal mean (left) and standard deviation (right) of the OOD feature. The ID distribution mean of the feature is assumed to have mean $\mu_0=-0.5$ and the OOD distribution is assumed to have mean $\mu_1-0.5$. The standard deviation of both distributions are assumed to have $\sigma=0.5$. The IB hyper-parameters are $\alpha=1.0, \beta=10$.}
    \label{fig:Gaussian_Low_Reg_Mean_Std}
\end{figure}

For this study, we will assume the assumptions of Section~\ref{subsec:numerical_optimization}, i.e., that the OOD features are element-wise independent and that the OOD feature is Gaussian, i.e., $p(\zt|z)\sim \mathcal N(\mu(z),\sigma_c(z))$. We will further assume that the ID distribution is Gaussian, i.e., $p(z|0)\sim \mathcal N(\mu_0,\sigma)$. We make this assumption for simplicity and that features in network layers can be approximated well with a Gaussian, as evidenced empirically in \citep{xu2023vravariationalrectifiedactivation}. We will study three OOD distributions next: Gaussian, Laplacian and a distribution we propose based on the Inverse Gaussian.

\begin{figure*}
    \centering
    \begin{subfigure}{0.27\textwidth}
        \centering
        \includegraphics[width=\textwidth] {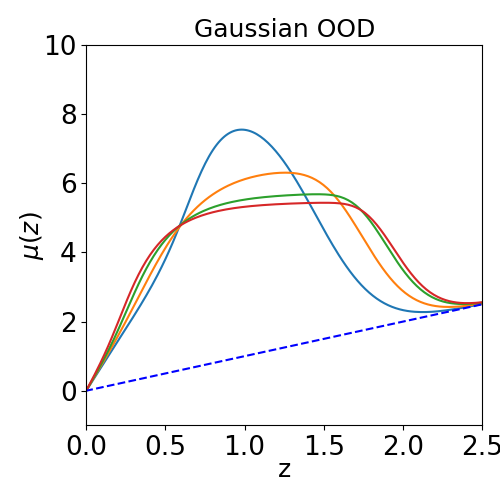}\\
        \vspace{-0.25cm}
        \caption{}
    \end{subfigure}
    \begin{subfigure}{0.27\textwidth}
        \centering
        \includegraphics[width=\textwidth]{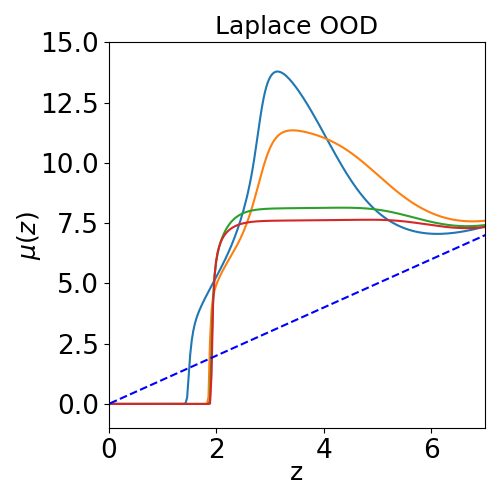}\\
        \vspace{-0.25cm}
        \caption{}
    \end{subfigure}
        \begin{subfigure}{0.27\textwidth}
        \centering
        \includegraphics[width=\textwidth] {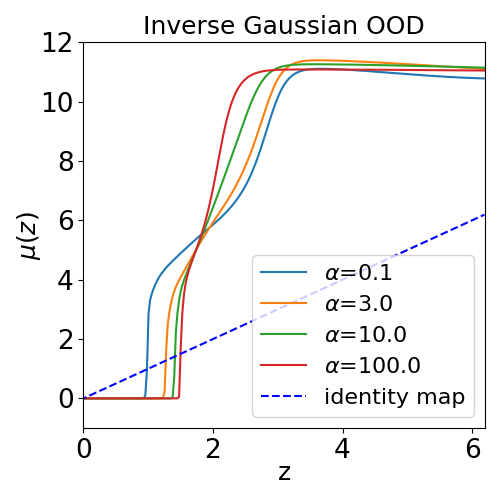}\\
        \vspace{-0.25cm}
        \caption{}
    \end{subfigure}
    \vspace{-0.5cm}
    \caption{The mean of the Optimal OOD Gaussian Feature under the Gaussian (left), Laplace (middle) and Inverse Gaussian (right) OOD distributions for varying weights on the Information Bottleneck, $\alpha$. For all plots, $\beta=10$. For the Gaussian case, $p(z|0)\sim \mathcal N(-0.5,0.5)$ and $p(z|1)\sim \mathcal N(0.5,0.5)$. For the Laplace case, $p(z|0)\sim \mathcal N(0,0.66)$ and $p(z|1)\sim Lap(0,1)$. In the Inverse Gaussian case, $p(z|0)\sim \mathcal{N}(0,0.66)$ and $p(z|1)\sim IG(d(z); 3.3,15)$. For visualization purpose, we only show the positive part.}
    \label{fig:OODfeature_Laplace_IG}
\end{figure*}

{\bf Gaussian OOD Distribution}: First, we study the case of a Gaussian for the OOD distribution, as its the most common distribution to study. Let $p(z|1)\sim \mathcal N(\mu_1, \sigma)$. For illustration, we choose $\mu_0=-0.5, \mu_1=0.5$ and $\sigma=0.5$ and $\alpha=1.0, \beta=10$. The resulting converged result of the optimization for $\mu$ and $\sigma_c$ is shown in Figure~\ref{fig:Gaussian_Low_Reg_Mean_Std} (positive part). No feature shaping would mean that $\mu$ is the identity map, and $\sigma_c=0$; this function is plotted in dashed blue. Notice that the optimal mean value is not the identity. The mean indicates that the feature has positive slope for small values of $|z|$ (similar to \citep{sun2021react}) and negative slope for large values of $|z|$ (similar to \citep{zhao2024towards}). In Appendix~\ref{app:plots_dist}, we show that under different distribution parameters, one can get negative values for small $|z|$ as in \citep{zhao2024towards}. Interestingly, the optimal standard deviation $\sigma_c(z)$ is non-zero, indicating randomness in this case is beneficial in terms of the loss. In fact, in all of our simulations across distributions and their hyperparameters, we have observed non-zero standard deviation.

\begin{figure}[h!]
    \centering
    \includegraphics[width=\linewidth]{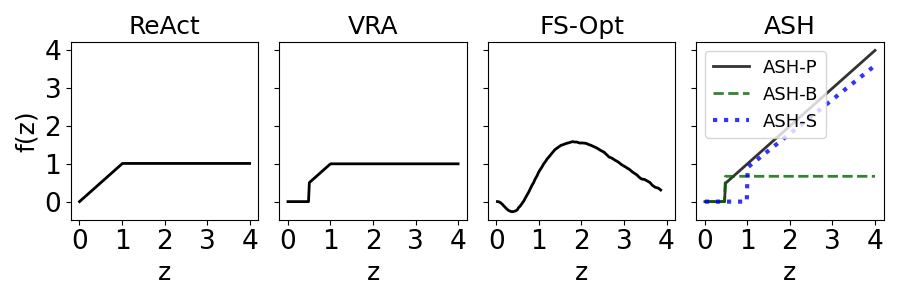}\\
    \vspace{-0.6cm}
    \caption{Plot of existing feature shaping functions from SoA methods: ReAct \citep{sun2021react}, VRA \citep{xu2023vravariationalrectifiedactivation}, FS-Opt \citep{zhao2024towards}, and variants of ASH \citep{djurisic2022ash}. Features generated with ResNet50 penultimate layer embeddings on ImageNet1k benchmark~\citep{xu2023vravariationalrectifiedactivation}. 
    }
    \label{fig:existing_features}
    \vspace{-0.2cm}
\end{figure}

In Figure~\ref{fig:OODfeature_Laplace_IG}(a), we show the effects of the Information Bottleneck weight $\alpha$. The impact of $\beta$ on the shape is studied in Appendix~\ref{app:plots_beta}.
For $\alpha$ larger (higher regularization), the mean of the feature becomes flat for $|z|$ large, similar to clipping that is used in popular methods \citep{sun2021react,xu2023vravariationalrectifiedactivation}.\footnote{\footnotesize Note our approach does not assume energy scoring like SoA, which changes the scale of the features. However, our approach uncovers and explains similar shapes irrespective of the score.} See Figure~\ref{fig:existing_features} for a plot of existing feature shaping methods. Even under the simplifying Gaussian assumptions, we see that the our shaping functions have properties of existing methods.


{\bf Laplacian OOD Distribution}: Next, we consider the Laplacian distribution for the OOD distribution, i.e., $p(z|1) = \frac{1}{2b}\exp{(-|z-\mu_1|/b)}$. The intuition for choosing this distribution is that it has a heavier tail than the Gaussian, and thus, is better able to model outliers, and it seems reasonable OOD data would be considered outliers. We show the result of the mean of the feature in Figure~\ref{fig:OODfeature_Laplace_IG}(b). We notice that when $|z|$ is small, the mean OOD feature is zero, which indicates a suppression of low values (this is used in VRA \citep{xu2023vravariationalrectifiedactivation} and ASH \citep{djurisic2022ash}). Note that this is consistent across $\alpha$ values, larger values increases the suppression region. We also see that large values of $|z|$ are being clipped or suppressed (with larger $\alpha$) approaching a zero slope. The jump discontinuity is also present in VRA and ASH. There also appears to be a positively sloped linear function for intermediate values of $|z|$, similar to VRA.

{\bf Inverse Gaussian OOD Distribution}: Next, we consider a distribution that may be a distribution that generically holds for OOD data and can be used in the absence of prior information of the OOD distribution. If the ID distribution is Gaussian, we can formulate a distribution that has high probability outside the domain that the ID distribution has high probability. To this end, one can consider a variant of the Inverse Gaussian defined as follows. Let $d(z) = |z-\mu_0|/\sigma$ where $\mu_0,\sigma$ are the mean and standard deviation of the ID distribution. This is a distance to the ID distribution. We would like the OOD distribution to have high probability when $d(z)$ is large, and thus we consider $p(z|1) \sim IG(d(z); \mu_1, \lambda)$ where IG denotes the inverse Gaussian distribution:
\begin{equation}
    p_{IG}(x; \mu, \lambda) = 
    \sqrt{\frac{\lambda}{2\pi x^3}} 
    \exp\left(
        -\frac{\lambda(x-\mu)^2}{2\mu^2x}
    \right),
\end{equation}
which is plotted in Appendix~\ref{app:InverseGaussian}. Note that there is some overlap of this distribution with the ID Gaussian. As noted in Figure~\ref{fig:OODfeature_Laplace_IG}(c), the Inverse Gaussian distribution results in a qualitatively similar OOD feature compared to the Laplacian distribution: suppression of small $|z|$ values and clipping/flattening of large $|z|$ values and a positively sloped linear function for intermediate values of $|z|$. For $\alpha$ large we have flattening similar to clipping and $\alpha$ smaller results in a negative slope similar to the other distributions.

{\bf Summary of Key Observations}: Clipping as done in ReAct seems to be a universal property across all the OOD distributions for large regularization.  In Appendix~\ref{app:gaussian_noise_activations}, we show empirically that for noiser OOD datasets larger regularization is beneficial, and so the clipping mitigates noise, which is noted in \citep{sun2021react}. Next, the OOD distributions that are heavier tailed result in suppression (zeroing out) of low $|z|$ values. This is consistent with the VRA and ASH methods. All distributions yield a positively sloped region for intermediate values of $|z|$. Our results suggest ReAct and FS-OPT may be operating under an implicit Gaussian OOD assumption for high regularization (ReAct) and low regularization (FS-OPT). VRA and ASH seem to implicitly assume heavier tailed OOD distributions.


\section{A New Shaping Function Motivated By Our Theory}
\label{sec: plf_motivation}
In this section, we show how our theory for feature construction developed in the previous sections can lead to a new shaping function that is both theoretically motivated and can achieve state of the art results as we show in the next section. The optimal mean shaping functions (from Gaussian, Laplace and Inverse Gaussian OOD distributions) can all be approximated by a few parameter piecewise linear function family. We propose to use such a piecewise linear function, which we abbreviate as PLF, as a new shaping function. We construct PLF to approximate any of the mean functions of the optimal feature under any of the three OOD distributions studied in the previous section. As many prominent existing feature shaping functions can be considered as arising from our framework under particular distributional assumptions as shown in the previous section, PLF can operate optimally under more general OOD distributions than those existing methods because it approximates any optimal shape under any of distributional assumptions (implicitly) assumed by those methods.

\begin{figure}
    \centering    \includegraphics[width=1\linewidth]{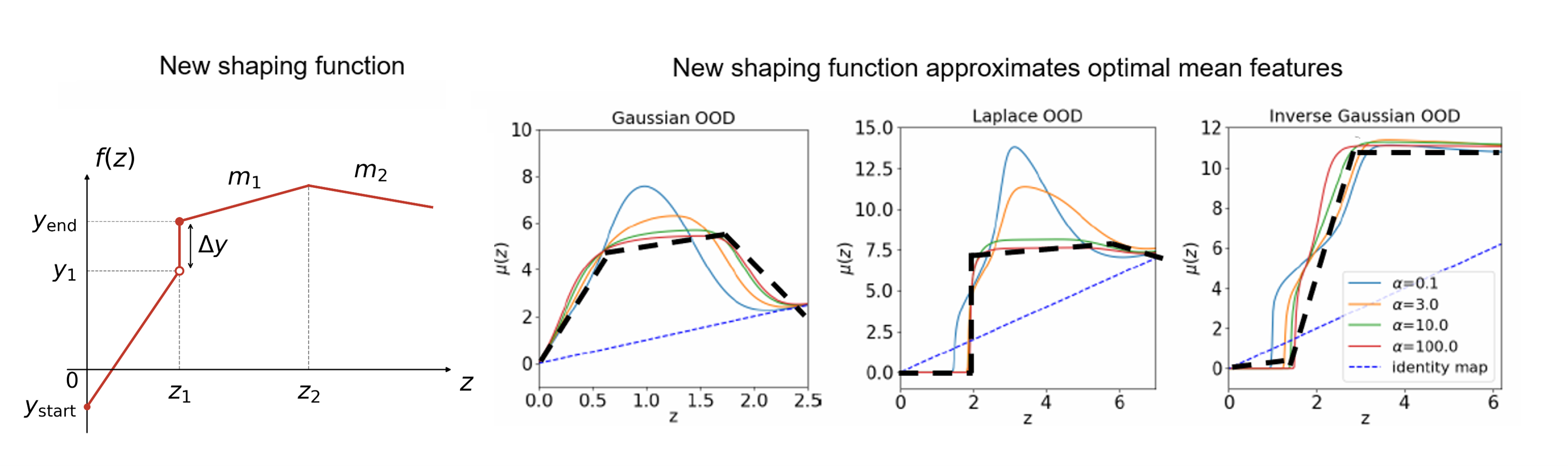}\\
    \caption{Our new piece-wise linear family of functions (left) that approximately encompasses the mean value of our OOD feature shaping functions across OOD distributions examined in this paper (three right plots).}
    \label{fig:piecewise_linear}
\end{figure}

We now define PLF. Figure~\ref{fig:piecewise_linear} shows a plot of the PLF family and its ability to approximate the mean shapes of optimal OOD features under the three OOD distributions discussed in the previous section. In each of the three right-hand panels of Figure~\ref{fig:piecewise_linear}, the dashed black line shows the PLF approximation, which approximates the optimal mean feature shape $\mu$ under the different OOD distributional assumptions. PLF is a seven-parameter piecewise linear shaping function, defined as
\begin{equation}
    f(z) =
    \begin{cases}
    y_{\text{start}} + \dfrac{y_1 - y_{\text{start}}}{z_1}\, z & 0 \leq z < z_1 \\[6pt]
    y_{\text{end}} + m_1 (z - z_1) & z_1 \leq z < z_2 \\[4pt]
    y_{\text{end}} + m_1(z_2-z_1) + m_2 (z - z_2) & z \geq z_2
    \end{cases}
    \label{eq:plf}
\end{equation}
where $z_1$ and $z_2$ are breakpoints and $m_1, m_2$ are the slopes of the second and third segments. The first segment rises linearly from $y_{\text{start}}$ to the value $y_{\text{end}} = y_1 + \Delta y$ at $z_1$, and the second segment begins at $y_{\text{end}}$. Note that $y_{\text{start}}$ may take on negative values as the optimal shapes can also take on negative values for certain distributions (Appendix~\ref{app:plots_dist}).
The function can be discontinuous at $z_1$ with jump size $\Delta y \geq 0$, and continuous at $z_2$ by construction, since the third segment starts where the second ends. 

The parameters of PLF must be tuned for practical use, which we discuss this in detail in Section~\ref{sec:implementation}, and Appendix~\ref{app: PLF_Bopt_settings}. The break-points $x_1$ and $x_2$ could be tuned directly, but their appropriate ranges depend on the range of the network features. Therefore, a direct parameterization is difficult to transfer across architectures, datasets, and feature dimensions. To obtain a range-adaptive formulation, we instead parameterize the breakpoints using quantiles of the empirical ID feature distribution, as described next.
Let $Z_{\mathrm{ID}}$ denote the scalar feature values obtained from the ID calibration data, and let $F_{|Z_{\mathrm{ID}}|}$ denote the empirical cumulative distribution function of their magnitudes. The breakpoints are defined as
\begin{equation}
z_1 = F^{-1}_{|Z_{\mathrm{ID}}|}(q_1), \qquad z_2 = F^{-1}_{|Z_{\mathrm{ID}}|}(q_2), \qquad q_2 = q_1 + \delta,
\label{eq:plf_quantiles}
\end{equation}
where $q_1 \in [0,1]$ determines the first transition point and $\delta > 0$ controls the separation between the two quantiles, subject to $q_1 + \delta \leq 1$. Because the breakpoints are now expressed as quantiles rather than raw feature values, they adapt automatically to the feature range of any given model and any given feature dimension. Under this formulation, the paramters of PLF are
\begin{equation}
\theta = \left( y_{\mathrm{start}}, y_{\mathrm{end}}, \Delta y, q_1, \delta, m_1, m_2 \right).
\end{equation}

This family of shaping functions is well suited to the setting where the OOD distribution is unknown: by tuning $\theta$, PLF can implicitly assume the optimal shape of any of the three distributions considered above without committing to one in advance. 
In the experimental section we explore this family and show that it achieves state-of-the-art results.


\section{Implementation of New OOD Detection Using Optimal Features}
\label{sec:implementation}


In this section, we provide the implementation details for our new approaches to OOD detection, using the simplifying assumptions presented in Section~\ref{sec:optimization}. We provide the details for two cases where the ID/OOD distributions are known and unknown. In the latter case, we apply the piecewise family of feature shaping derived in the previous section (Figure~\ref{fig:piecewise_linear}). We assume that a validation set of ID and OOD data is available (as in existing literature, see e.g., \citep{zhang2024openoodv15enhancedbenchmark}) and these choices are given in our experiments section. A trained neural network is also provided. The network feature vectors and their ID/OOD label for the validation set are $\{\mathbf z^i, y^i\}$. Consistent with our simplifying assumptions and literature, each component of the network feature $z$ is processed independently and for this paper, they will be processed by the same shaping function $\mu$.

\subsection{Off-line Training} 
\label{sec:bayesopt}

Under the case that the forms of the ID and OOD distributions are known, the hyperparameters of the distributions are estimated from the validation set (rasterizing the vector data). Using the fitted distributions, Algorithm~\ref{alg:numerical_gradient_descent} is run to compute the optimal $\mu^{\ast}, \sigma_c^{\ast}$. In the case that the distributions are unknown, we assume that the feature shape fits the piecewise family in the previous section (i.e., the OOD distributions are one of Gaussian, Laplacian or IG). The hyper-parameters for the piecewise family are tuned by e.g., maximizing the area under Receiver Operating Characteristics (ROC) metric on the validation set - this gives the optimal shaping function $\mu^{\ast}$.


The seven parameters of the PLF shaping function are tuned using black-box optimization via Bayesian optimization (BO)~\citep{frazier2018tutorialbayesianoptimization} by maximizing AUROC over a tuning set comprising validation ID and OOD data. We use a Gaussian process (GP) surrogate to model this objective as a function of the parameter vector, implemented via \texttt{scikit-optimize}. At each iteration, the GP posterior mean and uncertainty over previously evaluated configurations are used to select the next candidate through an acquisition function that balances exploiting configurations known to perform well against exploring regions of the search space that remain uncertain; the true objective is then evaluated at this candidate and the GP is updated. The search begins with an initial phase of randomly sampled configurations before switching to acquisition-guided proposals, with candidates evaluated in small batches to keep the surrogate model current. For each candidate, the quantile-based breakpoints are computed from the empirical distribution of absolute in-distribution features, the resulting shaping function is applied to both ID and OOD features, and the shaped features are passed through the network's frozen classifier head to obtain an energy score, from which the AUROC of ID-versus-OOD separation is computed as the optimization objective.

See Algorithm~\ref{alg:offline_training} for the algorithm for offline training.

\begin{algorithm}[h!]
\caption{Offline training and calibration}
\label{alg:offline_training}
\begin{algorithmic}[1]

\Require Frozen classifier head $h$; labeled validation features
$\mathcal{D}_{\mathrm{val}}
=\{(\mathbf{z}^i,y^i)\}_{i=1}^{n_{\mathrm{val}}}$
\Ensure Optimal shaping function $\mu^\ast$ and decision threshold $\tau^\ast$

\If{the ID and OOD distribution forms are known}
    \State Estimate the distribution parameters from
    $\mathcal{D}_{\mathrm{val}}$
    \State Compute $\mu^\ast$ and $\sigma_c^\ast$ using
    Algorithm~\ref{alg:numerical_gradient_descent}; retain $\mu^\ast$ as the deterministic shaping function
\Else
    \State Initialize Bayesian optimization over the parameters $\theta$
    of the piecewise shaping function $f_\theta$ in
    Eq.~\eqref{eq:plf}
    \For{$t=1,\ldots,T$}
        \State Propose candidate parameters $\theta_t$
        \State Determine the breakpoints from the corresponding
        quantiles of $|Z_{\mathrm{ID}}|$
        \State Compute shaped features
        $\widetilde{\mathbf{z}}^i=f_{\theta_t}(\mathbf{z}^i)$
        \State Compute the energy scores of the shaped features       $s_{i,t}=\mathcal{E}\!\left(h(\widetilde{\mathbf{z}}^i)\right)$
        \State Evaluate the validation objective (AUROC) and update the
        Bayesian-optimization surrogate
    \EndFor
    \State Set
    $\theta^\ast\gets\arg\max_t\operatorname{AUROC}_t$
    and $\mu^\ast\gets f_{\theta^\ast}$
\EndIf

\State Compute
$s_i=\mathcal{E}\!\left(h(\mu^\ast(\mathbf{z}^i))\right)$
for all validation features
\State Calibrate the decision threshold $\tau^\ast$ using the validation scores
\State \Return $\mu^\ast$ and $\tau^\ast$

\end{algorithmic}
\end{algorithm}

\subsection{Online Operation}

During operation, the network feature $\bf z$ is extracted, and then shaped via the function $\mathbf{\tilde z} = \mu^{\ast}(\mathbf{z})=(\mu^{\ast}(z_1), \ldots, \mu^{\ast}(z_n))$.  Subsequently, $\mathbf{\tilde z}$ is input to a scoring function (e.g., in this paper, energy score \citep{liu2021energy}), which is then thresholded to produce the ID/OOD label.  See Algorithm~\ref{alg:online_inference}.

\begin{algorithm}[h!]
\caption{Online ID/OOD inference}
\label{alg:online_inference}
\begin{algorithmic}[1]

\Require Incoming input $\mathbf{x}$; trained network
$N=h\circ g$; offline-learned shaping function $\mu^\ast$;
offline-calibrated decision rule $\delta_{\tau^\ast}$
\Ensure Predicted label $\widehat{y}\in\{\mathrm{ID},\mathrm{OOD}\}$

\State Extract the network feature:
$\mathbf{z} \gets g(\mathbf{x})$
\State Apply the offline-learned shaping function component-wise:
\[
    \widetilde{\mathbf{z}}
    \gets \mu^\ast(\mathbf{z})
    =
    \bigl(
        \mu^\ast(z_1),\ldots,\mu^\ast(z_n)
    \bigr)
\]
\State Pass the shaped feature through the frozen classifier head:
$\widetilde{\mathbf{v}} \gets h(\widetilde{\mathbf{z}})$
\State Compute the energy score
$s \gets \mathcal{E}(\widetilde{\mathbf{v}})$
as in \citep{liu2021energy}
\State Apply the offline-calibrated decision rule:
$\widehat{y} \gets \delta_{\tau^\ast}(s)$
\State \Return $\widehat{y}$

\end{algorithmic}
\end{algorithm}


\section{Experiments}
\label{sec: experiments}
We validate our theory by comparing our new shaping function to SoA for OOD detection on standard benchmarks. First, we test the performance on semantic shifts using two benchmarks, one used by prominent feature shaping methods and the OpenOOD benchmark \citep{zhang2024openoodv15enhancedbenchmark} broadly used for evaluating OOD methods. Second, we compare our method to state-of-the-art OOD detection methods on covariate shifts using the BROAD dataset \citep{guille2024expecting} and OpenOOD Covariate Shift datasets.

\subsection{Semantic Shift Benchmarking}

We compare the performance of our method on two benchmarks, one considered in the feature-shaping literature \citep{zhang2024learning} and the OpenOOD benchmark \citep{zhang2024openoodv15enhancedbenchmark}.

\subsubsection{Feature-Shaping Benchmark}

\textbf{Datasets and Model Architectures.}  
 We experiment with ResNet-50\citep{Resnet50}, MobileNet-v2~\citep{mobilenetv2}, vision transformers ViT-B-16 and ViT-L-16~\citep{dosovitskiy2021_VIT_B}  with ImageNet-1k~\citep{russakovsky2015imagenet} as ID data, and benchmark on the OOD datasets/methods used in \citep{zhao2024towards}.  For the ImageNet benchmark, we evaluate performance across eight OOD datasets: Species \citep{hendrycks2022_species}, iNaturalist \citep{vanhorn2018inaturalistspeciesclassificationdetection}, SUN \citep{SUN_data}, Places \citep{places}, OpenImage-O \citep{wang2022_OpenImageO}, ImageNet-O \citep{Hendrycks_2021_Imagenet-O}, Texture \citep{cimpoi_textures}, and MNIST \citep{mnist_deng}.
Moreover, we also experiment with CIFAR 10 and CIFAR 100 as ID data, for which we use a ViT-B-16 \citep{dosovitskiy2021_VIT_B} finetuned on CIFAR10/100 consistent with \citep{Fort}, and a MLP-Mixer-Nano model trained on CIFAR10/100 from scratch. We evaluate eight OOD datasets: TinyImageNet \citep{tiny_imagenet}, SVHN \citep{svhn}, Texture \citep{cimpoi_textures}, Places365 \citep{places}, LSUN-Cropped \citep{yu2016_lsun_cropped}, LSUN-Resized \citep{yu2016_lsun_cropped}, iSUN \citep{xu2015_isun}, and CIFAR100/ CIFAR10 (CIFAR 100 treated as OOD for CIFAR 10, and vice-versa).

We compare our results against the SoA methods across two categories - penultimate layer element-wise feature shaping approaches, which our theory currently applies to, and other methods. Penultimate layer feature shaping approaches involve element-wise feature shaping functions applied directly to the penultimate layer of the model before computing the energy score for OOD detection. Approaches in this category are: Energy~\citep{liu2021energy}, ReAct~\citep{sun2021react}, BFAct~\citep{Kong_BFAct}, VRA-P~\citep{xu2023vravariationalrectifiedactivation} and FS-OPT~\citep{zhao2024towards}. The second category, which are not directly comparable to our approach because they may not involve feature shaping or additions to feature matching, but are included for reference, are softmax-based confidence scoring (MSP~\citep{msp_hendrycks}), input perturbation and temperature scaling (ODIN~\citep{odin_liang}), intermediate-layer shaping and subsequent processing by following layers (ASH-P, ASH-B, ASH-S~\citep{djurisic2022ash}), or weight sparsification (DICE~\citep{sun2022dice}).


As in ReAct~\citep{sun2021react}, for ImageNet-1k benchmarks we use a validation set comprising the validation split of ImageNet-1k as ID data, and Gaussian noise images as OOD data, generated by sampling from $\mathcal N (0,1)$ for each pixel location, to tune the hyperparameters of our piecewise linear activation shaping function. For CIFAR 10/100 benchmarks, following ODIN~\citep{odin_liang}, we employ a random subset of the iSUN dataset~\citep{xu2015_isun} as validation OOD data for our hyperparameter tuning. As ID validation data for CIFAR10/100 we use the test splits of the corresponding datasets. The hyperparameters are optimized using Bayesian optimization~\citep{frazier2018tutorialbayesianoptimization}, by minimizing the FPR95 metric on the validation set. 

\textbf{Metrics.}  We utilize two standard evaluation metrics, following \citep{sun2021react,zhao2024towards}: FPR95 - the false positive rate when the true positive rate is 95\% (abbreviated as FP), and the area under the ROC curve (AU).

\textbf{Results. } The results on the ImageNet-1k benchmarks (Table~\ref{table_resnet}) and the CIFAR 10/100 benchmarks (Table~\ref{tab:cifar_results}) demonstrate that our approach achieves state-of-the-art performance among comparable feature-shaping methods in the previously mentioned category of methods. Specifically, when compared to pointwise feature-shaping techniques such as ReAct, BFAct, VRA-P, and FS-OPT, our method consistently outperforms these approaches, yielding the best overall results in this category.

\begin{table*}[ht]
\centering
\small
\begin{tabular}{cllcccccccc}
\hline 
& &  \multicolumn{2}{c}{ ResNet50 } & \multicolumn{2}{c}{ MobileNetV2 } & \multicolumn{2}{c}{ ViT-B-16 } & \multicolumn{2}{c}{ ViT-L-16 } \\
\cline{3-10} 
&  Method & FP $\downarrow$ & AU $\uparrow$ & FP $\downarrow$ & AU $\uparrow$ & FP $\downarrow$ & AU $\uparrow$ & FP $\downarrow$ & AU $\uparrow$ \\
\hline
\multirow{6}{*}{\rotatebox[origin=c]{90}{\shortstack{\scriptsize{Element-wise} \\ \scriptsize{Feature Shaping}}}}
 & Energy & 60.97 & 81.01 & 61.40 & 82.83 & 73.96 & 67.65 & 74.89 & 70.11 \\
&  ReAct & 42.29 & 86.54 & 54.19 & 85.37 & 73.82 & 76.86 & 76.16 & 81.07 \\
&  BFAct & 43.87 & 86.01 & 52.87 & 85.78 & 77.64 & 80.16 & 84.02 & 81.12 \\
&  VRA-P & 37.97 & 88.58 & 49.98 & 86.83 & 98.39 & 35.66 & 99.58 & 16.70 \\
&  FS-OPT & 39.75 & 88.56 & 51.77 & 86.62 & 69.52 & \textbf{82.66} & 72.17 & 83.23 \\
&  Ours & \textbf{35.82} & \textbf{89.36} & \textbf{46.97} & \textbf{87.49} & \textbf{67.73} & 81.06 & \textbf{66.67} & \textbf{83.92} \\
\hline
\multirow{6}{*}{\rotatebox[origin=c]{90}{\small{{Other methods}}}} &  MSP & 69.30 & 76.26 & 72.58 & 77.41 & \underline{69.84} & \underline{77.40} & 70.59 & \underline{78.40} \\
&  ODIN & 61.56 & 80.92 & 62.91 & 82.64 & 69.25 & 72.60 & \underline{70.35} & 74.51 \\
&  ASH-P & 55.30 & 83.00 & 59.41 & 83.84 & 99.36 & 21.17 & 99.18 & 20.27 \\
&  ASH-B & 35.97 & 88.62 & \underline{43.59} & \underline{88.28} & 94.87 & 46.68 & 93.72 & 38.95 \\
&  ASH-S & \underline{34.70} & \underline{90.25} & 43.84 & 88.24 & 99.48 & 18.52 & 99.42 & 18.61 \\
& DICE & 45.32 & 83.64 & 49.33 & 84.63 & 89.68 & 71.32 & 72.38 & 67.08 \\
\hline
\end{tabular}
\caption{ImageNet-1k Benchmarking Results. Lower FP is better, higher AU is better. Bold values highlight the best results among element-wise feature shaping methods. Underlined values indicate best results among other methods. }
\label{table_resnet}
\end{table*}

While ASH variants marginally outperform our method in some cases, it is important to note that ASH employs a fundamentally different approach. ASH modifies activations through intermediate layer pruning and rescaling of features, which are then routed back into the network for further processing, and thereby is not an element-wise feature shaping approach. Our theory currently does not address this case. 

For the vision transformers ViT-B-16 and ViT-L-16, our method achieves the lowest FP among all competing methods, providing evidence of generalization across different architectures. Overall, our results demonstrate that our feature shaping is highly competitive with the latest SoA, along with providing a theoretical explanation. 

\setlength{\tabcolsep}{1.5pt}
\begin{table}
\scriptsize{
    \centering
    \renewcommand{\arraystretch}{1.1}
    \begin{tabular}{c l c c c c c c c c}
        \toprule
        & & \multicolumn{4}{c}{\textbf{CIFAR10}} & \multicolumn{4}{c}{\textbf{CIFAR100}} \\
        \cmidrule(lr){3-6} \cmidrule(lr){7-10}
        & & \multicolumn{2}{c}{DenseNet101} & \multicolumn{2}{c}{MLP-N} & \multicolumn{2}{c}{DenseNet101} & \multicolumn{2}{c}{MLP-N} \\
        &  Method  & FP$\downarrow$ & AU$\uparrow$ & FP$\downarrow$ & AU$\uparrow$ & FP$\downarrow$ & AU$\uparrow$ & FP$\downarrow$ & AU$\uparrow$ \\
        \midrule
        \multirow{6}{*}{\rotatebox[origin=c]{90}{\shortstack{\scriptsize{Element-wise} \\ \scriptsize{Feature Shaping}}}}
        & Energy & 31.72 & 93.51 & 63.95 & \textbf{82.84} & 70.80 & 80.22  & 79.90 & 75.30 \\
        & ReAct & 82.00 & 76.46 & 64.34 & 81.85 & 77.00 & 78.30 & 79.99 & 75.87 \\
        & BFact & 84.40 & 74.39 & 78.02 & 72.68 & 80.27 & 73.36 & 80.05 & \textbf{76.58} \\
        & VRA-P & 38.41 & 92.77 & 100.00 & 65.95 & 67.75 & \textbf{82.72} & 87.19 & 66.03 \\
        & FS-OPT & 28.90 & 94.12 & 83.87 & 71.83 & 65.20 & 82.39 & 81.33 & 74.67 \\
        & Ours & \textbf{24.08} &\textbf{95.26} & \textbf{62.67} & 81.67 & \textbf{64.15} & 82.00 & \textbf{78.33} & 73.53 \\
        \midrule
        \multirow{6}{*}{\rotatebox[origin=c]{90}{\small{{Other methods}}}}
        & MSP & 52.66 & 91.42 &  67.01 & 82.86 & 80.40 & 74.75 & 83.97 & 73.07 \\
        & ODIN & 32.84 & 91.94 & 69.20 & 69.53 & 62.03 & 82.57 & \underline{78.71} & 66.47 \\
        & MLS & 31.93 & 93.51 &  \underline{64.50} & \underline{82.97} & 70.71 & 80.18 & 82.41 & \underline{74.52} \\
        & ASH-P & 29.39 & 93.98 & 84.39 & 66.93 & 68.21 & 81.11 & 86.73 & 65.27 \\
        & ASH-B & 28.21 & 94.27 & 93.93 & 53.00 & 57.45 & 83.80 & 93.63 & 57.20 \\
        & ASH-S & \underline{23.93} & \underline{94.41} & 82.57 & 68.02 & \underline{52.41} & \underline{84.65} & 89.39 & 59.63 \\
        & DICE & 29.67 & 93.27 &  96.64 & 52.17 & 59.56 & 82.26 & 95.78 & 44.48 \\
        \bottomrule
    \end{tabular}
    \caption{Performance comparison on CIFAR10/100 datasets. Lower FP is better, higher AU is better. Bold values highlight the best results among element-wise feature shaping methods. Underlined values indicate best results among other methods. }
    \label{tab:cifar_results}}
    \vspace{-0.5cm}
\end{table}

\subsubsection{OpenOOD Benchmark}

{\bf Datasets and Architectures}: We evaluate our method \footnote{Code: https://github.com/maryhadalittlecode/PLF-OpenOOD} using the protocols and data from the OpenOOD benchmark~\citep{zhang2024openoodv15enhancedbenchmark}, which includes four in-distribution (ID) datasets: CIFAR-10~\citep{krizhevsky2009learning}, CIFAR-100~\citep{krizhevsky2009cifar}, ImageNet-200, and ImageNet-1K~\citep{imagenet_paper}, each paired with predefined near- and far-OOD datasets. For CIFAR-10, the near-OOD datasets are CIFAR-100 and Tiny ImageNet, while the far-OOD datasets are MNIST~\citep{mnist_deng}, SVHN~\citep{svhn}, Textures~\citep{cimpoi_textures}, and Places365~\citep{places}. For CIFAR-100, the near-OOD datasets are CIFAR-10 and Tiny ImageNet, with the same far-OOD datasets as CIFAR-10. For ImageNet-1K, the near-OOD datasets are SSB-hard~\citep{vaze2022openset} and NINCO~\citep{bitterwolf2023outfixingimagenetoutofdistribution}, and the far-OOD datasets are iNaturalist~\citep{vanhorn2018inaturalistspeciesclassificationdetection}, Textures, and OpenImage-O~\citep{wang2022_OpenImageO}. The ImageNet-200 benchmark is a 200-class subset of ImageNet-1K that uses the same near- and far-OOD datasets. Following the OpenOOD protocol, we use a ResNet-18~\citep{he2015deepresiduallearningimage} backbone for CIFAR-10, CIFAR-100, and ImageNet-200, and a ResNet-50~\citep{Resnet50} backbone for ImageNet-1K.

{\bf Methods Compared}: 
We compare against the post-hoc inference methods in OpenOOD benchmark. These methods are applied at the inference phase and by default assume that the classifier is trained with the standard cross-entropy loss. We use network checkpoints provided in OpenOOD for evaluation.

{\bf Settings}: PLF is tuned using Bayesian optimization on fixed ID and OOD validation sets, strictly following the OpenOOD benchmark protocol. For CIFAR-10, we use 1,000 samples held out from the official test set as ID tuning data and 1,000 images spanning 20 categories from Tiny ImageNet as OOD tuning data. We use the same protocol for CIFAR-100. For ImageNet-1K, the tuning sets comprise 5,000 images from the ImageNet validation set as ID data and 1,763 images from OpenImage-O as OOD data. For ImageNet-200, we use the corresponding 1,000-image subset of the ImageNet validation set as ID tuning data and the same 1,763 OpenImage-O images as OOD tuning data. 

{\bf Results}:  Table~\ref{tab:ood_detection_results} compares our method (PLF) against both feature-shaping methods and a broad range of post-hoc OOD detectors on four standard OpenOOD benchmarks spanning CIFAR-10, CIFAR-100, ImageNet-200, and ImageNet-1K.  We evaluate based the Near-OOD, Far-OOD and AVG-OOD metrics, which are the average AUROC performance on near OOD datasets, far OOD datasets, and all OOD datasets, respectively. Overall, PLF demonstrates the most consistent performance across datasets. Among feature-shaping approaches, PLF achieves the highest Avg-OOD AUROC on three of the four benchmarks and is competitive on CIFAR-100, where all feature-shaping methods perform similarly. Importantly, PLF ranks within the top three methods on every benchmark (unlike any other method) and achieves the best overall ranking on the challenging ImageNet-1K benchmark. The improvements are observed across both Near-OOD and Far-OOD settings, indicating that the proposed feature shaping improves robustness without introducing a trade-off between the two types of OOD. Compared with existing feature-shaping approaches (Energy, ReAct, and VRA-P), PLF consistently improves OOD detection accuracy, while remaining competitive with the strongest distance-based methods such as KNN and RMDS.

\setlength{\tabcolsep}{1.5pt}
\begin{table*}[htbp]
\centering
\scriptsize
\renewcommand{\arraystretch}{1.1}
\resizebox{\textwidth}{!}{%
\begin{tabular}{c l|cccc|cccc|cccc|cccc}

\toprule
& \multirow{2}{*}{Method}
& \multicolumn{4}{c|}{CIFAR-10 (ResNet-18)}
& \multicolumn{4}{c|}{CIFAR-100 (ResNet-18)}
& \multicolumn{4}{c|}{ImageNet-200 (ResNet-18)}
& \multicolumn{4}{c}{ImageNet-1K (ResNet-50)} \\
\cmidrule(lr){3-6}
\cmidrule(lr){7-10}
\cmidrule(lr){11-14}
\cmidrule(lr){15-18}
& 
& Near-OOD & Far-OOD & Avg-OOD & Ranking
& Near-OOD & Far-OOD & Avg-OOD & Ranking
& Near-OOD & Far-OOD & Avg-OOD & Ranking
& Near-OOD & Far-OOD & Avg-OOD & Ranking \\
\midrule

\multirow{4}{*}{\rotatebox[origin=c]{90}{\shortstack{\scriptsize{Feature } \\ \scriptsize{Shaping}}}}
& Energy & $87.58_{\pm0.46}$ & $91.21_{\pm0.92}$ & $90.00_{\pm0.99}$ & 6 & $80.91_{\pm0.08}$ & $79.77_{\pm0.61}$ & $80.15_{\pm0.73}$ & 6 & $82.50_{\pm0.05}$ & $90.86_{\pm0.21}$ & $88.07_{\pm0.21}$ & 8 & 75.89 & 89.47 & 84.03 & 8 \\
& ReAct & $87.11_{\pm0.61}$ & $90.42_{\pm1.41}$ & $89.31_{\pm1.63}$ & 9 & \bm{$80.77_{\pm0.05}$} & $80.39_{\pm0.49}$ & $80.52_{\pm0.54}$ & 5 &  $81.87_{\pm0.98}$ & $92.31_{\pm0.56}$ & $88.83_{\pm0.71}$ & 4 & 77.38 & 93.67 & 87.15 & 4 \\
& VRA-P & $87.11_{\pm0.75}$ & $90.41_{\pm1.73}$ & $89.31_{\pm1.99}$ & 9 & \bm{$80.77_{\pm0.07}$} & $80.42_{\pm0.60}$ & $80.53_{\pm0.66}$ & 4&  $81.96_{\pm1.18}$ & $92.46_{\pm0.73}$ & $88.26_{\pm0.92}$ & 7& 77.91 & 94.36 & 87.77 & 3 \\
& Ours & \bm{$89.47_{\pm0.89}$} & \bm{$92.36_{\pm1.09}$} & \bm{$91.40_{\pm1.48}$} & 3 & $80.75_{\pm0.19}$ & \bm{$80.53_{\pm1.38}$} & \bm{$80.60_{\pm1.24}$} &3 &  \bm{$83.28_{\pm1.11}$} & \bm{$93.82_{\pm0.85}$} & \bm{$89.60_{\pm0.97}$} & 2 &  \textbf{78.15} &  \textbf{96.24} & $\bm{89.00}$ & 1 \\

\midrule

\multirow{15}{*}{\rotatebox[origin=c]{90}{\scriptsize{Other methods}}}
& MSP & $88.03_{\pm0.25}$ & $90.73_{\pm0.43}$ & $89.83_{\pm0.61}$ & 8 & $80.27_{\pm0.11}$ & $77.76_{\pm0.44}$ & $78.60_{\pm0.67}$ & 13 &  $83.34_{\pm0.06}$ & $90.13_{\pm0.09}$ & $87.87_{\pm0.11}$ & 10 & 76.02 & 85.23 & 81.55 & 17 \\
& ODIN & $82.87_{\pm1.85}$ & $87.96_{\pm0.61}$ & $86.26_{\pm1.66}$ & 13 & $79.90_{\pm0.11}$ & $79.28_{\pm0.21}$ & $79.49_{\pm0.61}$ & 10 &  $80.27_{\pm0.08}$ & $91.71_{\pm0.19}$ & $87.90_{\pm0.19}$ & 9 & 74.75 & 89.47 & 83.58 & 10 \\
& MLS & $87.52_{\pm0.47}$ & $91.10_{\pm0.89}$ & $89.90_{\pm0.97}$ & 7 & \underline{$81.05_{\pm0.07}$} & $79.67_{\pm0.57}$ & $80.14_{\pm0.70}$ & 7 &  $82.90_{\pm0.04}$ & $91.11_{\pm0.19}$ & $88.38_{\pm0.19}$ & 6 & 76.46 & 89.57 & 84.33 & 6 \\
& ASH & $75.27_{\pm1.04}$ & $78.49_{\pm2.58}$ & $77.42_{\pm3.22}$ & 17 & $78.20_{\pm0.15}$ & $80.58_{\pm0.66}$ & $79.79_{\pm0.54}$ & 9 &  $82.38_{\pm0.19}$ & \underline{$93.90_{\pm0.27}$} & \underline{$90.06_{\pm0.27}$} & 1 & \underline{78.17} & \underline{95.74} & \underline{88.71} & 2 \\
& DICE & $78.34_{\pm0.79}$ & $84.23_{\pm1.89}$ & $82.27_{\pm2.80}$ & 15 & $79.38_{\pm0.23}$ & $80.01_{\pm0.18}$ & $79.80_{\pm0.92}$ & 8 &  $81.78_{\pm0.14}$ & $90.80_{\pm0.31}$ & $87.79_{\pm0.32}$ & 11 & 73.07 & 90.95 & 83.80 & 9 \\
& SHE & $81.54_{\pm0.51}$ & $85.32_{\pm1.43}$ & $84.06_{\pm1.61}$ & 14 & $78.95_{\pm0.18}$ & $76.92_{\pm1.16}$ & $77.59_{\pm1.21}$ & 15 &  $80.18_{\pm0.25}$ & $89.81_{\pm0.61}$ & $86.60_{\pm0.55}$ & 14 & 73.78 & 90.92 & 84.07 & 7 \\
& RankFeat & $79.46_{\pm2.52}$ & $75.87_{\pm5.06}$ & $77.07_{\pm4.54}$ & 18 & $61.88_{\pm1.28}$ & $67.10_{\pm1.42}$ & $65.36_{\pm2.12}$ & 21 &  $56.92_{\pm1.59}$ & $38.22_{\pm3.85}$ & $44.45_{\pm3.17}$ & 22 & 50.99 & 53.93 & 52.75 & 21 \\
& KLM & $79.19_{\pm0.80}$ & $82.68_{\pm0.21}$ & $81.52_{\pm0.91}$ & 16 & $76.56_{\pm0.25}$ & $76.24_{\pm0.52}$ & $76.35_{\pm0.71}$ & 16 &  $80.76_{\pm0.08}$ & $88.53_{\pm0.11}$ & $85.94_{\pm0.15}$ & 14 & 76.64 & 87.60 & 83.22 & 14 \\
& VIM & $88.68_{\pm0.28}$ & \underline{$93.48_{\pm0.24}$} & $91.88_{\pm0.37}$ & 2 & $74.98_{\pm0.13}$ & $81.70_{\pm0.62}$ & $79.46_{\pm1.08}$ & 11 &  $78.68_{\pm0.24}$ & $91.26_{\pm0.19}$ & $87.07_{\pm0.23}$ & 12 & 72.08 & 92.68 & 84.44 & 5 \\
& KNN & \underline{$90.64_{\pm0.20}$} & $92.96_{\pm0.14}$ & \underline{$92.19_{\pm0.26}$} & 1 & $80.18_{\pm0.15}$ & $82.40_{\pm0.17}$ & $81.66_{\pm0.72}$ & 2 &  $81.57_{\pm0.17}$ & $93.16_{\pm0.22}$ & $89.29_{\pm0.21}$ & 3 & 71.10 & 90.18 & 82.55 & 16 \\
& OpenGAN & $53.71_{\pm7.68}$ & $54.61_{\pm15.51}$ & $54.31_{\pm15.18}$& 22 & $65.98_{\pm1.26}$ & $67.88_{\pm7.16}$ & $67.25_{\pm6.03}$ & 18 &  $59.79_{\pm3.39}$ & $73.15_{\pm4.07}$ & $68.69_{\pm3.89}$ & 20 & N/A & N/A & N/A & N/A \\
& GradNorm & $54.90_{\pm0.98}$ & $57.55_{\pm3.22}$ & $56.67_{\pm4.19}$ & 21 & $70.13_{\pm0.47}$ & $69.14_{\pm1.05}$ & $69.47_{\pm1.19}$ & 17 &  $72.75_{\pm0.48}$ & $84.26_{\pm0.87}$ & $80.43_{\pm0.88}$ & 17 & 72.96 & 90.25 & 83.34 & 13 \\
& MDS & $84.20_{\pm2.40}$ & $89.72_{\pm1.36}$ & $87.88_{\pm1.88}$ & 12 & $58.69_{\pm0.09}$ & $69.39_{\pm1.39}$ & $65.82_{\pm1.48}$ & 20 &  $61.93_{\pm0.51}$ & $74.72_{\pm0.26}$ & $70.46_{\pm0.42}$ & 18 & 55.44 & 74.25 & 66.72 & 19  \\
& MDSEns & $60.43_{\pm0.26}$ & $73.90_{\pm0.27}$ & $69.41_{\pm0.36}$ & 19 & $46.31_{\pm0.24}$ & $66.00_{\pm0.69}$ & $59.44_{\pm0.80}$ & 22 &  $54.32_{\pm0.24}$ & $69.27_{\pm0.57}$ & $64.29_{\pm0.51}$ & 21 & 49.67 & 67.52 & 60.38 & 20 \\
& RMDS & $89.80_{\pm0.28}$ & $92.20_{\pm0.21}$ & $91.40_{\pm0.34}$ & 3 & $80.15_{\pm0.11}$ & \underline{$82.92_{\pm0.42}$} & \underline{$82.00_{\pm0.80}$} & 1&  $82.57_{\pm0.25}$ & $88.06_{\pm0.34}$ & $86.23_{\pm0.32}$ & 15& 76.99 & 86.38 & 82.63 & 15 \\
& Gram & $58.66_{\pm4.83}$ & $71.73_{\pm3.20}$ & $67.38_{\pm4.69}$ & 20 & $51.66_{\pm0.77}$ & $73.36_{\pm1.08}$ & $66.13_{\pm1.57}$ & 19 &  $67.67_{\pm1.07}$ & $71.19_{\pm0.24}$ & $70.02_{\pm0.59}$ & 19 & 61.70 & 79.71 & 72.50 & 18 \\
& OpenMax & $87.62_{\pm0.29}$ & $89.62_{\pm0.19}$ & $88.95_{\pm0.39}$ &11 & $76.41_{\pm0.25}$ & $79.48_{\pm0.41}$ & $78.46_{\pm0.65}$ & 14&  $80.27_{\pm0.10}$ & $90.20_{\pm0.17}$ & $86.89_{\pm0.16}$ & 13 & 74.77 & 89.26 & 83.46 & 11 \\
& TempScale & $88.09_{\pm0.31}$ & $90.97_{\pm0.52}$ & $90.01_{\pm0.70}$ & 5 & $80.90_{\pm0.07}$ & $78.74_{\pm0.51}$ & $79.46_{\pm0.67}$ & 11 &  \underline{$83.69_{\pm0.04}$} & $90.82_{\pm0.09}$ & $88.45_{\pm0.10}$ & 5 & 77.14 & 87.56 & 83.39 & 12 \\
\bottomrule
\end{tabular}}

\caption{OOD detection performance on the OpenOOD benchmark using ResNet-18 backbones for CIFAR-10, CIFAR-100, and ImageNet-200, and a ResNet-50 backbone for ImageNet-1K. Near-OOD and Far-OOD report the mean AUROC over the corresponding near- and far-distribution-shift test sets. For CIFAR-10, Avg-OOD is averaged over CIFAR-100, TIN, MNIST, SVHN, Textures, and Places365; for CIFAR-100, it is averaged over CIFAR-10, TIN, MNIST, SVHN, Textures, and Places365. For ImageNet-200 and ImageNet-1K, Avg-OOD is averaged over SSB-hard, NINCO, iNaturalist, Textures, and OpenImage-O. Results are reported as mean $\pm$ standard deviation where available, and higher AUROC indicates better performance. Rankings are computed separately for each benchmark using the Avg-OOD mean across all methods. Bold values indicate the best results among feature-shaping methods, while underlined values indicate the best results among the other methods.}
\label{tab:ood_detection_results}
\end{table*}

\subsection{Covariate Shift Benchmarking}

Covariate shifts can result in OOD and cause the network to produce unreliable results, and hence it's important for OOD detectors to be able to detect covariate shift \cite{guille2024expecting}.  We test our method against other state of the art in this scenario. To the best of our knowledge, this is the first comprehensive benchmarking of state-of-the-art post-hoc OOD detection methods of OpenOOD on covariate shifts. 

{\bf Datasets and Architectures}: 
We evaluate on diverse covariate shifts from OpenOOD~\citep{zhang2024openoodv15enhancedbenchmark} together with the BROAD benchmark~\citep{guille2024expecting}. From OpenOOD, we use ImageNet-V2 (resampling bias)~\citep{imagenet-v2}, ImageNet-R (style shifts)~\citep{imagenet-r}, and ImageNet-C (corruptions)~\citep{imagenet-c}. The BROAD benchmark is specifically designed to capture diverse covariate shifts, comprising adversarial attacks (AutoAttack~\citep{autoattack} and Projected Gradient Descent (PGD)~\citep{madry2019deeplearningmodelsresistant}), synthetic images (Stable Diffusion~\citep{diffusion} and conditional BigGAN~\citep{cBigGAN}), and a multi-label dataset~\citep{guille2024expecting}. All experiments in this setting use ImageNet-1K~\citep{imagenet_paper} as the ID dataset with a ResNet-50~\citep{Resnet50} backbone.

{\bf Methods Compared}: 
We compare against the post-hoc inference methods in OpenOOD benchmark. These methods are applied at the inference phase and by default assume that the classifier is trained with the standard cross-entropy loss. We use network checkpoints provided in OpenOOD for evaluation.

{\bf Settings}: We used the tuned detectors using the standard protocols on the standard OpenOOD benchmark described in the previous sub-section (that is on the same validation data of semantic shifted data), and we evaluate on the covariate shifted data.

{\bf Results}: Table~\ref{tab:imagenet1k_covariate_auroc} shows the results of our method compared to state of the art. Note such covariate shifts may be OOD to the network and produce unreliable results, and hence should be detected. PLF achieves the highest average (over all datasets in the benchmarks) AUROC among all feature-shaping methods and ranks second overall across all evaluated methods, with comparable accuracy as the top-ranking method. Unlike competing approaches that specialize on only a subset of perturbations, PLF performs strongly across a diverse collection of shifts, obtaining the best feature-shaping performance on ImageNet-C, AutoAttack, PGD, and Cond.BigGAN while remaining competitive on the remaining benchmarks. In particular, PLF substantially improves robustness to adversarial and generative distribution shifts, which are among the most challenging settings for post-hoc OOD detection. These results demonstrate that the learned feature shaping transfers beyond semantic OOD benchmarks to a broad spectrum of real-world covariate shifts.

\setlength{\tabcolsep}{2.0pt}
\begin{table*}[htbp]
\centering
\scriptsize
\renewcommand{\arraystretch}{1.15}
\resizebox{\textwidth}{!}{%
\begin{tabular}{c l|*{13}{c}}
\toprule
& Method
& \makecell{ImageNet-C}
& \makecell{ImageNet-V2}
& \makecell{ImageNet-R}
& \makecell{Auto-Attack}
& PGD
& \makecell{Stable-Diff}
& \makecell{Cond.BigGAN}
& \makecell{Multi-Label}
& Avg.
& Rank
\\
\midrule

\multirow{4}{*}
{\rotatebox[origin=c]{90}{
\fontsize{5}{5.5}\selectfont
{\shortstack{\scriptsize{Feature } \\ \scriptsize{Shaping}}}}
}
& Energy & 82.54 & 58.63 & \textbf{87.03} & 84.95 & 57.45 & 73.10 & 42.39 & 66.20 & 69.04 & 8 \\
& ReAct  &  82.33 & 58.48 & 85.98 & 84.55 & 59.22 &  \textbf{73.79} & 38.85 &  \textbf{69.49} & 69.08 & 7\\
& VRA-P  & 82.78 & \textbf{58.53} & 85.99 & 84.92 & 59.51 & 73.71 & 40.19 & 68.88 & 69.31 & 5 \\
& Ours  &  \textbf{84.70} & 58.09 & 86.27 & \textbf{86.70} & \textbf{62.09} & 72.03 & \textbf{53.97} & 65.99 & \textbf{71.23} & 2 \\

\midrule

\multirow{18}{*}{\rotatebox[origin=c]{90}{\scriptsize{Other methods}}}
& MSP  & 78.54 & 58.54 & 80.51 & 82.40 & 62.54 & 69.85 & 41.42 & \underline{70.14} & 67.99 & 10 \\
& ODIN  & 76.77 & 57.60 & 85.47 & 79.10 & 49.42 & \underline{78.56} & 44.34 & 69.37 & 67.58 & 13 \\
& MLS  & 82.38 & \underline{58.77} & 86.70 & 84.98 & 57.74 & 73.17 & 42.18 & 67.40 & 69.16 & 6 \\
& ASH  & 83.91 & 58.29 & 85.21 & 86.25 & 60.37 & 70.30 & 54.12 & 65.09 & 70.44 & 3 \\
& DICE  & 82.28 & 56.98 & 82.85 & 83.98 & 55.12 & 67.58 & 50.94 & 62.75 & 67.81 & 12 \\
& SHE & 80.82 & 55.35 & 76.08 & 82.35 & 55.22 & 63.15 & 61.42 & 56.80 & 66.39 & 16  \\
& RankFeat & 61.89 & 51.16 & 62.95 & 58.10 & 46.86 & 63.52 & 46.49 & 46.71 & 54.71 & 19 \\
& KLM  & 77.99 & 57.58 & 81.62 & 81.96 & 62.35 & 69.92 & 41.94 & 67.26  & 67.58 & 14\\
& VIM & 84.85 & 57.33 & \underline{87.94} & 85.81 & 62.68 & 76.13 & 38.30 & 65.25 & 69.79 & 4 \\
& KNN  & 84.95 & 56.44 & 87.64 & 86.01 & 64.93 & 73.02 & 37.86 & 53.91 & 68.09 & 9 \\
& GradNorm  & 80.11 & 54.38 & 69.66 & 81.16 & 53.92 & 55.42 & \underline{75.60} & 51.12 & 65.17 & 17 \\
& MDS & 69.12 & 51.72 & 69.78 & 68.24 & 63.61 & 66.17 & 32.85 & 52.35  & 59.23 & 18 \\
& MDSEns & 46.70 & 41.40 & 60.63 & 37.86 & 40.46 & 66.15 & 28.81 & 41.46 & 45.43 & 21\\
& RMDS & 79.75 & 58.70 & 81.73 & 80.75 & \underline{65.14} & 73.66 & 27.52 & 65.30  & 66.56 & 15 \\
& Gram& \underline{88.04} & 55.56 & 81.82 & 85.03 & 58.77 & 79.49 & 31.52 & 62.58 & 67.85 & 11\\
& OpenMax  & 67.04 & 41.81 & 70.34 & 68.25 & 43.66 & 55.09 & 26.98 & 45.75 & 52.36 & 20\\
& TempScale  & 86.29 & 58.36 & 83.81 & \underline{87.94} & 63.41 & 70.60 & 61.56 & 61.89 & \underline{71.73} & 1 \\

\bottomrule
\end{tabular}%
}%
\caption{AUROC results on the ImageNet-1K covariate-shift benchmark using a ResNet-50 backbone. Avg. denotes the mean AUROC across the eight covariate-shift datasets, and Rank is computed over all methods according to Avg., with higher values indicating better performance. Bold values highlight the best results among feature-shaping methods, while underlined values indicate the best results among other methods.}
\label{tab:imagenet1k_covariate_auroc}
\end{table*}

\subsection{Results Summary}

\textbf{Overall Results on Semantic and Covariate Shifts}: Taken together, Tables~\ref{tab:ood_detection_results} and \ref{tab:imagenet1k_covariate_auroc} demonstrate that our method PLF consistently improves OOD detection across both semantic and covariate distribution shifts. Unlike existing methods that exhibit strong benchmark dependence, PLF remains among the top-performing approaches across all evaluation settings, suggesting that the proposed feature shaping captures a more general and transferable representation for OOD detection.  In fact, PLF is in the top three of all 21 methods evaluated on OpenOOD and BROAD benchmarks, unlike any other method.

\textbf{Computational Time}: The inference cost of our feature shaping method is on the order of microseconds for a 256×256×3 image, using PyTorch on an NVIDIA A100-80GB GPU. This is comparable to other piecewise linear shaping approaches such as ReAct and VRA.

\section{Conclusion and Future Work}

We have presented a novel theory for OOD feature construction. OOD features were computed as the optimizer of a loss functional consisting of a term maximizing the KL distance between resulting features under ID and OOD distributions and a regularization based on the Information Bottleneck. We have related the optimal features to several element-wise OOD shaping functions that are used in existing practice, offering a theoretical explanation and suggesting underlying distributional assumptions made in these often empirically motivated approaches. Our theory was shown to lead to a new shaping function (PLF), which was constructed specifically to operate under more general distributional assumptions than other feature shaping methods.  Extensive benchmarking across several benchmarks testing both semantic and covariate shift shows that our new method out-performs existing feature shaping approaches. More broadly PLF ranks in the top three of 21 evaluated state-of-the-art among all OOD detection methods, including both feature-shaping and non-feature shaping approaches, on all datasets/architectures. No other method performs consistently in the top three.

There are several areas for future investigation. Firstly, we have developed the concept of random features, whose mean value models OOD shaping functions, but only exploited the mean value algorithmically. Future work will aim to exploit the OOD feature distribution. Secondly, our theory so far only explains the OOD feature and not the score function. We wish to incorporate scores into our theory. Finally, we wish to explore more general vector shaping functions.




\appendix

\section{Analysis of Loss Function in 1D Gaussian Case}
\label{app:1DGaussianLoss}

\begin{figure}
    \centering
    \includegraphics[width=0.9\linewidth]{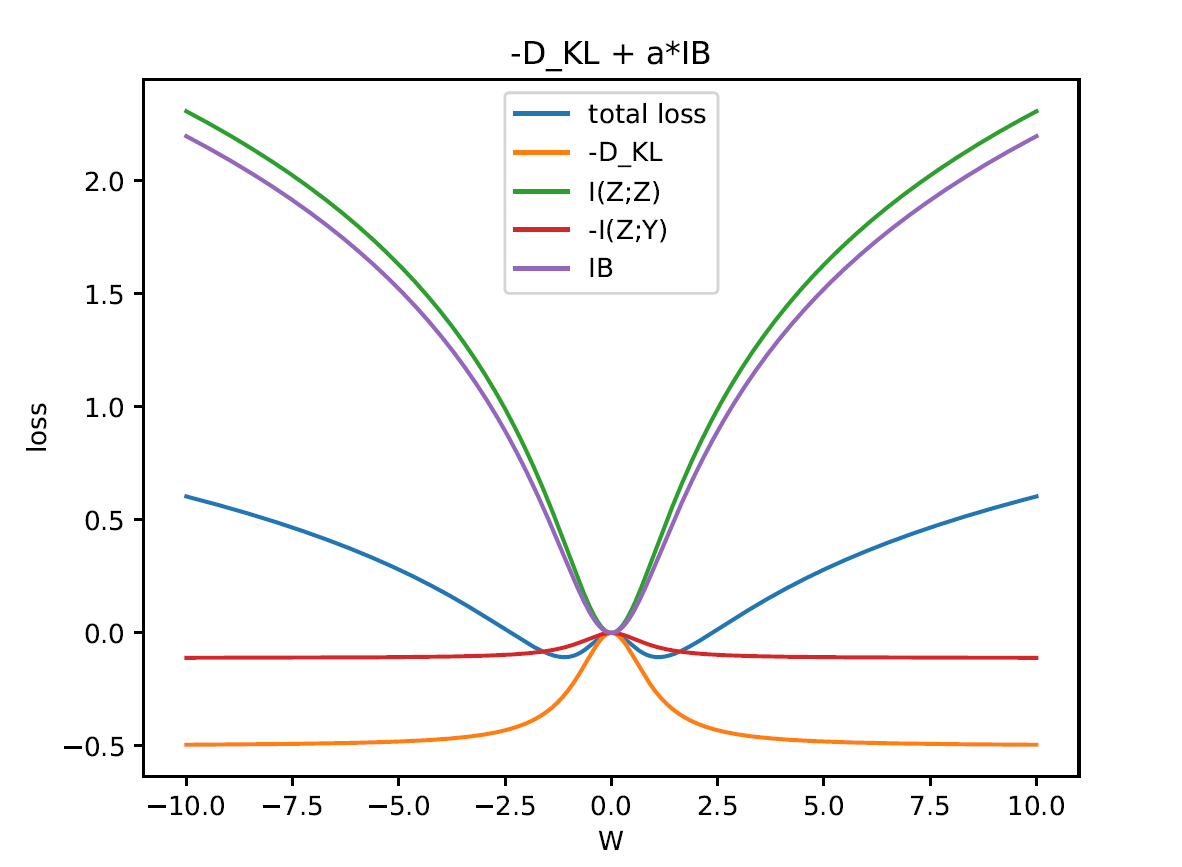}
    \caption{Loss function for the 1D Gaussian case.  Note $\mu_0-\mu_1=1$, $\sigma=\sigma_c=1$, and $\beta=1$.  Note $L=D_{KL}+\alpha \mathrm{IB}$ where $\alpha=0.5$.
    }
    \label{fig:loss_1d_plot}
\end{figure}

We analytically study the optimization of our loss functional in the case where the ID/OOD distributions of $z$ are Gaussian, i.e., $p(z|y) \sim \mathcal{N}(\mu_y, \sigma_y), \, y\in \{0,1\}$, and the feature is a Gaussian with mean being a linear function and the standard deviation constant, i.e., $p(\zt|z) \sim \mathcal{N}(Wz+b, \sigma_c)$ where $W,b\in \R$.   The $W,b$ are parameters that are to be optimized, which specify the OOD feature. This is a relaxation of deterministic shaping function being a linear function. The loss function shape and various components of the loss are shown in Figure~\ref{fig:loss_1d_plot}. The plot shows the loss terms versus $W$; as will be shown below the loss does not depend on $b$. It shows that the separation between feature distributions (KL term) increases as $|W|\to\infty$. On the otherhand, the information bottleneck term increases as $|W|\to 0$. Thus, these terms compete with each other and the optimal solution is well-defined at a finite $|W|>0$. This simple example suggests that the loss functional is well defined (i.e., has a finite optimum). Notice that without the IB term, there is no optimal value of $W$.

Next, we derive the components of the loss in analytic form. Based on the assumptions specified in the previous paragraph, the formulas for the assumed probabilities are:
\begin{align}
    p(z|y) &=  \frac{1}{\sqrt{2\pi}\sigma_y} 
    \exp{\left(-\frac{1}{2\sigma_y^2}(z-\mu_y)^2 \right)} \\
    p(\tilde z|z) &= \frac{1}{\sqrt{2\pi}\sigma_c} 
    \exp{\left( -\frac{(\tilde z - Wz-b)^2}{2\sigma_c^2} \right)}.
\end{align}
Under these assumptions, we can calculate the feature distributions:
\begin{align}
    p(\tilde z|y) &=
    \frac{1}{\sqrt{2\pi(\sigma_c^2+W^2\sigma_y^2)}} 
    \exp{\left(-\frac{1}{2(\sigma_c^2+W^2\sigma_y^2)}(\tilde z-W\mu_y-b)^2 \right)},
\end{align}
and note that the joint distribution is
\begin{equation}
    p(\tilde z,z|y) \sim N(\mu_{\tilde z,z}, \Sigma_{\tilde z,z}), \quad
    \mu_{\tilde z, z}
    = 
    \left(
        \begin{array}{c}
            W\mu_y + b \\
            \mu_y
        \end{array}
    \right),
    \quad
    \Sigma_{z,\tilde z} = 
    \left(
        \begin{array}{cc}
            \sigma_c^2 + W^2\sigma^2  & W\sigma^2 \\
            W\sigma^2 & \sigma^2
        \end{array}
    \right).
\end{equation}

Let us compute the loss of both the KL term and the information bottleneck under this case. First note the formula: if $p\sim \mathcal{N}(\hat \mu_1,\hat \sigma_1), q\sim \mathcal{N}(\hat \mu_2,\hat \sigma_2)$ then 
\begin{equation}
    D_{KL}(p||q) = 
    \log{\left(\frac{\hat \sigma_2}{\hat \sigma_1}\right)} + 
    \frac{\hat \sigma_1^2+(\hat \mu_1-\hat \mu_2)^2}{2\hat \sigma_2^2}- \frac 1 2.
\end{equation}
Choosing
\begin{align}
    \hat \mu_1 &= W\mu_1+b,\\
    \hat \mu_2 &= W\mu_0+b, \\
    \hat \sigma_1^2, \hat \sigma_2^2 &= \sigma_c^2+W^2\sigma^2,
\end{align}
we find that
\begin{equation}
    D_{KL}( p(\tilde z|Y=1)\, ||\, p(\tilde z|Y=0) ) = 
    \frac{\sigma_c^2+W^2\sigma^2+W^2(\mu_1-\mu_0)^2}{2(\sigma_c^2+W^2\sigma^2)} - \frac 1 2 = 
    \frac{W^2(\mu_1-\mu_0)^2}{2(\sigma_c^2+W^2\sigma^2)}.
\end{equation}

Let us now compute the information bottleneck term. Note the following result: if $X,Y \sim \mathcal{N}(\mu,\Sigma)$, where 
\begin{equation}
    \Sigma = 
    \left(
       \begin{array}{ll}
        \sigma_X^2 & \rho \sigma_{X}\sigma_{Y} \\
         \rho \sigma_{X}\sigma_{Y} & \sigma_Y^2
       \end{array}
    \right),
\end{equation}
then 
\begin{equation}
    I(X; Y) = -\frac 1 2 \log{(1-\rho^2)}.
\end{equation}
Note that
\begin{equation}
    \Sigma_{Z, \tilde Z} = 
    \left(
    \begin{array}{cc}
        \sigma_c^2 + W^2\sigma^2 & W \sigma^2 \\
        W \sigma^2 & \sigma^2
    \end{array}
    \right); 
\end{equation}
therefore,
\begin{equation}
    \rho_{\tilde Z, Z} = \frac{W\sigma}{\sqrt{\sigma_c^2+W^2\sigma^2}}
\end{equation}
Thus,
\begin{equation}
    I(\tilde Z; Z) = 
    -\frac 1 2 \log\left[
        1- \frac{W^2\sigma^2}{\sigma_c^2+W^2\sigma^2}\right] = 
        -\frac 1 2 
        \log{\left(  \frac{\sigma_c^2}{\sigma_c^2+W^2\sigma^2}  \right)}.
\end{equation}
Next, we compute
\begin{align}
    I(\tilde Z; Y) &=
    p(Y=1)\int p(\tilde z|Y=1) \log{\frac{p(\tilde z|Y=1)}{p(\tilde z)}} \ud \tilde z + 
    p(Y=0)\int p(\tilde z|Y=0) \log{\frac{p(\tilde z|Y=0)}{p(\tilde z)}} \ud \tilde z \\
    &= -p(Y=1)h(p(\tilde z|Y=1)) - p(Y=0)h(p(\tilde z|Y=0)) + h(p(\tilde z)) \\
    &= -\frac 1 2 \left[ \log{(2\pi(\sigma_c^2+W^2\sigma^2))} + 1\right] + h(p(\tilde z)),
\end{align}
where we used that
\begin{equation}
    h(\mathcal{N}(\mu,\sigma^2)) = \frac 1 2 \left[
        \log(2\pi\sigma^2)+1
    \right].
\end{equation}
Note that
\begin{align}
    p(\tilde z) &= p(Y=1)p(\tilde z|Y=1) + p(Y=0)p(\tilde z|Y=0) \\
    &= p(Y=1)G(\tilde z'; W\mu_0+b, \sigma_{\tilde z}) + p(Y=0) G(\tilde z'; W\mu_1+b, \sigma_{\tilde z}) \\
    &= \frac{1}{\sigma_{\tilde z}}
    \left[
    p(Y=1)G(z'; 0, 1) + p(Y=0)G(z'; \mu', 1)
    \right],
\end{align}
where 
\begin{equation}
    z' = \frac{\tilde z - W\mu_0-b}{\sigma_{\tilde z}}, \quad 
    \sigma_{\tilde z}^2 = \sigma_c^2 + W^2\sigma^2, \quad
    \mu' = \frac{W(\mu_1-\mu_0)}{\sigma_{\tilde z}}.
\end{equation}
Thus, 
\begin{align}
    h(p(\tilde z)) &= -\int p(\tilde z) \log{p(\tilde z)} \ud \tilde z \\
    &= -\sigma_{\tilde z} \int p(z')\log p(z') \ud z' \\
    &= 
    \int \tilde G(z') \log{\left[ \frac{1}{\sigma_{\tilde z}} \tilde G(z') \right]}
    \ud z' \\
    &= \log{\sigma_{\tilde z}} - 
    \int \tilde G(z')\log \tilde G(z') \ud z'\\
    &= \log{\sigma_{\tilde z}} + h(\tilde G),
\end{align}
where 
\begin{equation}
    \tilde G(z') = p(Y=1)G(z'; 0, 1) + p(Y=0)G(z'; \mu', 1).
\end{equation}
Therfore,
\begin{align}
    I(\tilde Z; Y) &= 
    -\frac 1 2 \left[ \log(2\pi\sigma_{\tilde z}^2) + 1 \right] + \log \sigma_{\tilde z} + h(\tilde G) \\
    &= 
    -\frac 1 2 \left[ \log(2\pi) + 1 \right] + h(\tilde G).
\end{align}
Therefore,
\begin{equation}
    \mathrm{IB}(p(\tilde z|z)) = 
    \log\left(\frac{\sigma_{\tilde z}}{\sigma_c}
    \right)
    +\frac{\beta}{2} \left[ \log(2\pi) + 1 \right] -\beta h(\tilde G).
\end{equation}
Finally,
\begin{equation}
    L(p(\tilde z|z)) = 
    -\frac{W^2(\mu_1-\mu_0)^2}{2\sigma_{\tilde z}^2}
    + 
    \log\left(\frac{\sigma_{\tilde z}}{\sigma_c}
    \right)
    +\frac{\beta}{2} \left[ \log(2\pi) + 1 \right] -\beta h(\tilde G),
\end{equation}
where
\begin{align}
    \tilde G(z) &= p(Y=1)G(z; 0, 1) + p(Y=0)G(z; \mu', 1) \\
    \sigma_{\tilde z}^2 &= \sigma_c^2 + W^2\sigma^2 \\
    \mu' &= \frac{W(\mu_0-\mu_1)}{\sigma_{\tilde z}}.
\end{align}
We show the plot of this loss function in Figure~\ref{fig:loss_1d_plot}.

\section{Simplifying the Loss With Independence Assumptions}
\label{app:simplification_loss}


We simplify our loss functional under independence assumptions. Specifically, we assume that
\begin{align}
    p(\zt|z) &= \prod_{i=1}^n p(\zt_i|z_i) \\
    p(z|y) &= \prod_{i=1}^n p(z_i|Y=y).
\end{align}
Note that
\begin{align}
    p(\zt|y) &= \int p(\zt|z)p(z|y) \ud z \\
    &= \int \prod_i p(\zt_i|z_i)p(z_i|y) \ud z_1 \ldots \ud z_n \\
    &= \prod_i \int p(\zt_i|z_i)p(z_i|y) \ud z_i \\
    &= \prod_i p(\tilde z_i|y).
\end{align}

Therefore,
\begin{align}
    D_{KL}[p(\zt|0)\,||\,p(\zt|1)] &=
    \int p(\zt|0)
    \log \prod_i \frac{p(\zt_i|0)}{p(\zt_i|1)} \ud \zt \\
    &= 
    \sum_i \int p(\zt|0) \log \frac{p(\zt_i|0)}{p(\zt_i|1)} \ud \zt \\
    &= 
    \sum_i D_{KL}[p(\zt_i|0)\,||\,p(\zt_i|1)],
\end{align}
and by a similar computation, we get that
\begin{align}
    I(\Zt; Z) &= \sum_i I(\Zt_i; Z_i) \\
    I(\Zt; Y) &= \sum_i I(\Zt_i; Y).
\end{align}
Thus, we see that
\begin{align}
    L(p(\zt|z)) = -D_KL[p(\zt|0)\,||\,p(\zt|1)] + 
    \alpha \mbox{IB}(p(\zt|z))  = 
    \sum_i L_i(p(\zt_i|z_i)),
\end{align}
where
\begin{align}
    L_i(p(\zt_i|z_i)) = 
    -D_{KL}[p(\zt_i|0)\,||\,p(\zt_i|1)] + 
    \alpha [ I(\Zt_i; Z_i) -\beta I(\Zt_i;Y) ].
\end{align}
Therefore, we just need to solve $n$ independent problems:
\begin{align}
    \argmin_{p(\zt_i|z_i)} L_i(p(\zt_i|z_i)), \quad i\in \{1,\ldots, n\}.
\end{align}

\section{Gradient of Loss Computations}
\label{app:gradients}

We review the definition of gradient of functionals, which are functions defined on functions, in order to compute gradients of our loss. First, we define the directional derivative. Let $\delta p(\zt|z)$ denote a perturbation of $p(\zt|z)$, which is a function of $\tilde z$ with integral zero so that $p(\zt|z) + \varepsilon \delta p(\zt|z)$ defines a valid probability (i.e., integrates to 1) for $\varepsilon$ small. The direction derivative is defined as
\begin{equation}
    \ud L(p(\zt|z)) \cdot \delta p(\zt|z) = \der{}{\varepsilon}
    \left.
    L[ p(\zt|z)+\varepsilon \delta p(\zt|z) ] 
    \right|_{\varepsilon=0}.
\end{equation}
The gradient $\nabla_{p(\zt|z)} L(p(\zt|z))$ is the perturbation of $p(\zt|z)$ that satisfies the relation:
\begin{equation}
    \ud L(p(\zt|z)) \cdot \delta p(\zt|z)  = 
    \int \nabla_{p(\zt|z)} L(p(\zt|z)) \cdot \delta p(\zt|z) \ud z.
\end{equation}

\subsection{KL Loss Gradient}

We look into the optimization of 
\begin{equation}
    \max_{p(\tilde z|z)} D_{KL}[ p(\tilde z|Y=1)\, ||\, p(\tilde z | Y=0) ],
\end{equation}
where $D_{KL}$ is the KL-divergence or relative entropy:
\begin{equation}
    D_{KL}[ p \, ||\, q ] = \int p(x)\log{\frac{p(x)}{q(x)}} \ud x,
\end{equation}
that is we would like to compute $\tilde Z$ such that the resulting distributions under in/out data are maximally separated.

We compute the optimizing conditions by computing the variation. First note the following:
\begin{equation}
    p(\tilde z | y ) = 
    \int p(\tilde z | z, y) p(z|y) \ud z = 
    \int p(\tilde z | z) p(z|y) \ud z,
\end{equation}
where the equality on the right hand side is by assumption - we do not want our feature $\tilde Z$ to be dependent on whether the data is OOD or not. Now we compute the variation:
\begin{align}
    \delta D_{KL} \cdot \delta p(\tilde z|z) &=
    \int \delta p(\tilde z|Y=1) \cdot \delta p(\tilde z|z) 
        \log{\left(\frac{p(\tilde z|Y=1)}{p(\tilde z|Y=0)}\right)} \\
    &+ 
        p(\tilde z | Y=0) \delta
        \left[
        \frac{p(\tilde z|Y=1)}{p(\tilde z|Y=0)}
        \right] \cdot \delta p(\tilde z|z) \ud \tilde z.
\end{align}
Let's compute 
\begin{align}
    \delta p(\tilde z|y) \cdot \delta p(\tilde z|z) = 
    \int \delta p(\tilde z|z) p(z|y) \ud z.
\end{align}
Therefore,
\begin{align}
    \delta
    \left[
    \frac{p(\tilde z|Y=1)}{p(\tilde z|Y=0)}
    \right]\cdot \delta p(\tilde z|z) &= 
    \frac{\delta p(\tilde z|Y=1) p(\tilde z|Y=0) -
        p(\tilde z|Y=1) \delta p(\tilde z|Y=0)
    }{p(\tilde z|Y=0)} \\
    &=
    \frac{
    \int \delta p(\tilde z|z) p(z|Y=1) \ud z \cdot p(\tilde z|Y=0) - 
    \int \delta p(\tilde z|z) p(z|Y=0) \ud z \cdot p(\tilde z|Y=1) }
    {p(\tilde z|Y=0)^2}.
\end{align}
Therefore, 
\begin{align}
p(\tilde z|Y=0)
\delta
\left[
\frac{p(\tilde z|Y=1)}{p(\tilde z|Y=0)}
\right]\cdot \delta p(\tilde z|z)
= 
\int \delta p(\tilde z|z) 
\left[ p(z|Y=1) - L(\tilde z) p(z|Y=0) \right] \ud z,
\end{align}
where
\begin{equation}
    l(\tilde z) = \frac{p(\tilde z|Y=1)}{p(\tilde z|Y=0)},
\end{equation}
is the likelihood ratio of the distributions of $\tilde Z$. Thus, 
\begin{align}
    \delta D_{KL} \cdot \delta p(\tilde z|z) &=
    \int \int \delta p(\tilde z|z)
    \left[
    \left(1 + \log l(\tilde z)\right)p(z|Y=1) - 
    l(\tilde z) p(z|Y=0)
    \right] \ud z \ud \tilde z,
\end{align}
where
\begin{equation}
    l(z) = \frac{p(z|Y=1)}{p(z|Y=0)} = 
    \frac{p_{in}(z)}{p_{out}(z)}.
\end{equation}
Therefore,
\begin{equation}
    \nabla_{p(\tilde z|z)} D_{KL} = 
    p_{out}(z) \cdot 
    \left[
        (1+\log l(\tilde z))l(z) - l(\tilde z)
    \right].
\end{equation}

\subsection{Information Bottleneck Gradient}
We consider the information bottleneck term:
\begin{equation}
    \mbox{IB}(p(\tilde z|z)) = I(Z; \tilde Z) - \beta I(\tilde Z; Y),
\end{equation}
where $I$ denotes mutual information. Note that
\begin{align}
    I(Z; \tilde Z) &= 
    \int p(z,\tilde z) \log{\frac{p(z,\tilde z)}{p(z)p(\tilde z)}} \ud \tilde z \ud z  = 
    \int p(z)p(\tilde z|z) \log{\frac{p(\tilde z|z)}{p(\tilde z)}} \ud \tilde z \ud z.
\end{align}
Also,
\begin{align}
    I(\tilde Z; Y) &= \sum_{y\in\{0,1\}} 
    \int p(\tilde z, y) \log{\frac{p(\tilde z,y)}{p(\tilde z)p(y)}} \ud \tilde z = 
    \sum_{y\in\{0,1\}} \int p(\tilde z|y) p(y) \log{\frac{p(\tilde z|y)}{p(\tilde z)}} \ud \tilde z.
\end{align}

Let us compute the variation of these terms:
\begin{align}
    \delta I(Z; \tilde Z) \cdot \delta p(\tilde z|z) &=
    \int p(z) \delta p(\tilde z|z) 
    \log{ \frac{p(\tilde z|z)}{p(\tilde z)} } + 
    p(z)
    \frac{\delta p(\tilde z|z)p(\tilde z) - p(\tilde z|z)\delta p(\tilde z) \cdot \delta p(\tilde z|z) }
    {p(\tilde z)} \ud \tilde z \ud z \\
    &=
    \int 
    \delta p(\tilde z|z) p(z) \left[ 1 + 
    \log{ \frac{p(\tilde z|z)}{p(\tilde z)} }
    \right] -
    \frac{p(\tilde z|z)p(z)}{p(\tilde z)}
    \int \delta p(\tilde z|z') p(z') \ud z' \ud \tilde z \ud z.
\end{align}
Let us evaluate the term after the minus sign:
\begin{align}
    \int \int \int \delta p(\tilde z|z') 
    \frac{p(\tilde z|z)p(z)}{p(\tilde z)} p(z') \ud z' \ud \tilde z \ud z &= 
    \int \int \delta p(\tilde z|z') \frac{p(z')}{p(\tilde z)} \int p(\tilde z|z)p(z)\ud z \ud \tilde z \ud z' \\
    &= \int\int \delta p(\tilde z|z') p(z') \ud \tilde z \ud z'.
\end{align}
Therefore,
\begin{align}
    \delta I(Z; \tilde Z) \cdot \delta p(\tilde z|z) &=
    \int \delta p(\tilde z|z) p(z) \log{\frac{p(\tilde z|z)}{p(\tilde z)}} \ud \tilde z \ud z, \\
    \nabla_{p(\tilde z|z)} I(Z; \tilde Z) &= p(z) 
    \log{\frac{p(\tilde z|z)}{p(\tilde z)}}.
\end{align}

Let us compute the variation of the second term in IB:
\begin{align}
    \delta I(\tilde Z; Y) &=
     \sum_{y\in\{0,1\}} \int \delta p(\tilde z|y) p(y) 
     \log{ \frac{p(\tilde z|y)}{p(\tilde z)} } + 
     p(y) 
     \frac{\delta p(\tilde z|y)p(\tilde z) - p(\tilde z|y)\delta p(\tilde z) }{p(\tilde z)}
     \ud z.
\end{align}
Note that 
\begin{align}
    \delta p(\tilde z) &= \int \delta p(\tilde z|z)p(z) \ud z\\
    \delta p(\tilde z|y) &= \int \delta p(\tilde z|z) p(z|y) \ud \tilde z.
\end{align}
Therefore,
\begin{align}
    \delta I(\tilde Z; Y) &=
    \sum_{y\in\{0,1\}} 
    \int\int \delta p(\tilde z|z)p(y) \left[ p(z|y) 
    \log{ \frac{p(\tilde z|y)}{p(\tilde z)} +
    p(z|y) - 
    \frac{p(\tilde z|y)p(z)}{p(\tilde z)}
    }
    \right] \ud \tilde z \ud z \\
    &= 
    \int\int \delta p(\tilde z|z) \sum_{y\in\{0,1\}} 
    p(y) \left[ p(z|y) 
    \log{ \frac{p(\tilde z|y)}{p(\tilde z)} +
    p(z|y) - 
    \frac{p(\tilde z|y)p(z)}{p(\tilde z)}
    }
    \right] \ud \tilde z \ud z \\
    &=
    \int\int \delta p(\tilde z|z) \sum_{y\in\{0,1\}} 
    p(y) p(z|y) 
    \log{ \frac{p(\tilde z|y)}{p(\tilde z)} } \ud \tilde z \ud z.
\end{align}
Therefore,
\begin{align}
    \nabla_{p(\tilde z|z)} \mbox{IB} &= 
    p(z) \log{\frac{p(\tilde z|z)}{p(\tilde z)}} - 
    \beta
    \sum_{y\in\{0,1\}} 
    p(y) p(z|y) 
    \log{ \frac{p(\tilde z|y)}{p(\tilde z)} } \\
    &=
    \sum_{y\in\{0,1\}} p(y) p(z|y) 
    \left[
    \log{\frac{p(\tilde z|z)}{p(\tilde z)}} - 
    \beta  \log{ \frac{p(\tilde z|y)}{p(\tilde z)} }
    \right].
\end{align}

\section{Plot of Inverse Gaussian Distribution}
\label{app:InverseGaussian}
\begin{figure}
    \centering
    \includegraphics[width=0.55\linewidth]{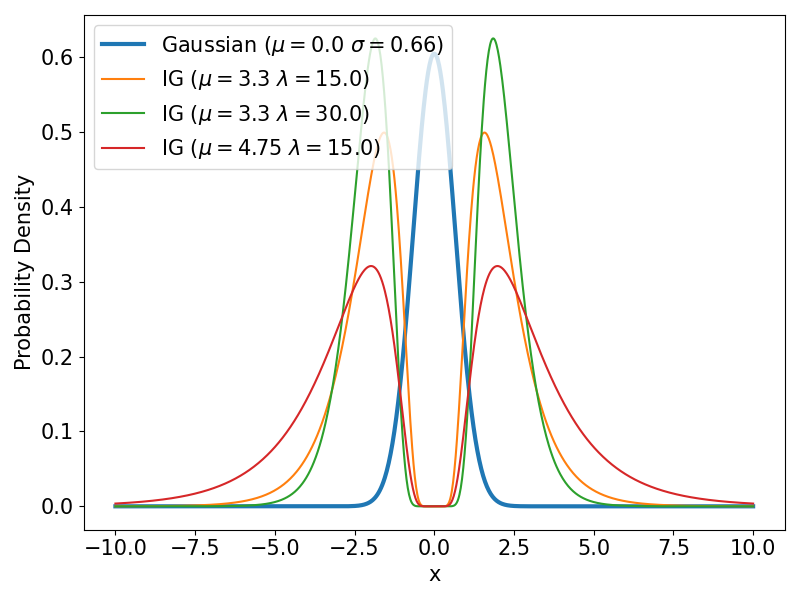}
    \caption{Plot of Inverse Gaussian (IG) distribution, $p(z|1)\sim IG(d(z);\mu,\lambda)$, under different parameters with a Gaussian (blue). Note that IG has high probability where the Gaussian does not.}
    \label{fig:inverse_gaussian}
\end{figure}

In Figure~\ref{fig:inverse_gaussian}, we show a plot of our Inverse Gaussian for various parameters along with a Gaussian.  Notice that the IG has mass complementary to the Gaussian, and thus represents a natural distribution for OOD if the ID is Gaussian. 

\section{A Study of Our Feature Shaping Over Parameters}

In this section, we perform an extended study of feature shaping as a function of parameters of the distributions and the $\beta$ parameter in the Information Bottleneck, extending the study in Section 4 of the main paper.

\subsection{Feature Shaping Versus $\beta$}
\label{app:plots_beta}
\begin{figure}
    \centering
    \includegraphics[clip,trim=0 10 10 10,width=0.32\linewidth]{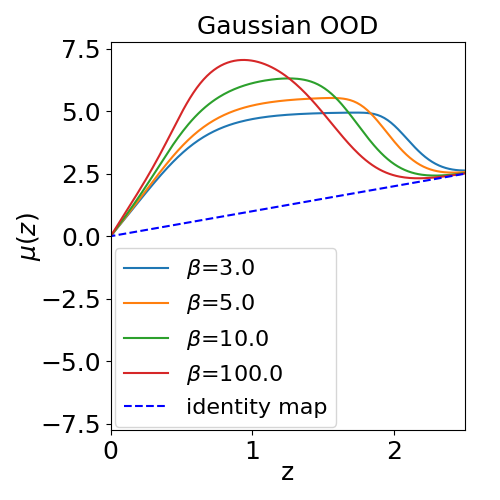}
    \includegraphics[clip,trim=0 10 10 10,width=0.32\linewidth]{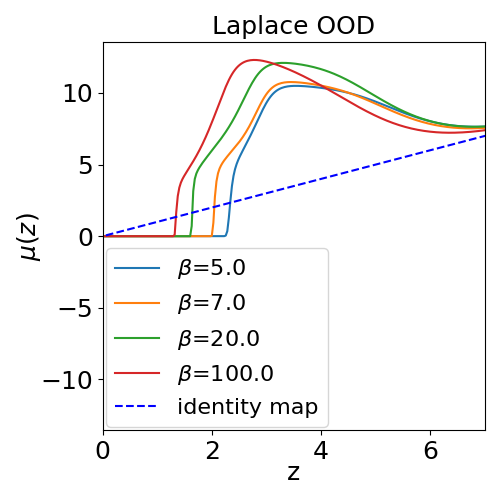}
    \includegraphics[clip,trim=0 10 10 10,width=0.32\linewidth]{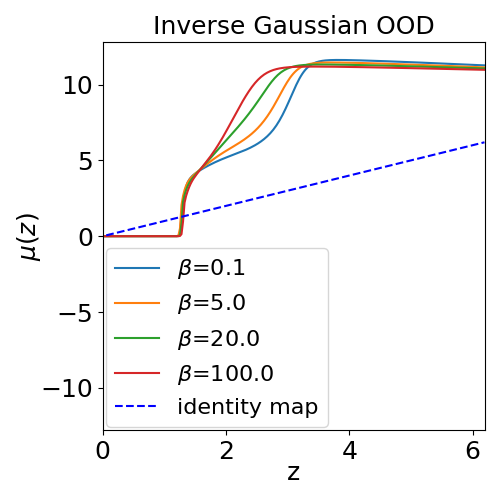}
    \caption{The mean of the OOD Gaussian Random Feature under the Gaussian (left), Laplace (middle) and Inverse Gaussian (right) distributions for the OOD distribution. Different curves on the same plot indicate differing weights on the $I(\tilde Z; Y)$ component of the Information Bottleneck term, $\beta$. The weight on the IB is fixed to $\alpha=3.0$. For the Gaussian case, $p(z|0)\sim \mathcal N(-0.5,0.5)$ and $p(z|1)\sim \mathcal N(0.5,0.5)$. For the Laplace case, $p(z|0)\sim \mathcal N(0,0.5)$ and $p(z|1)\sim Lap(0,1)$. In the Inverse Gaussian case, $p(z|0)\sim \mathcal{N}(0,0.5)$ and $p(z|1)\sim IG(d(z); 0.5,15)$. }
    \label{fig:OODfeature_Laplace_IG_beta}
\end{figure}

In Figure~\ref{fig:OODfeature_Laplace_IG_beta}, we explore how our feature shaping function varies as a function of $\beta$ for various distributions.

\subsection{Feature Shaping Versus Distribution Parameters}
\label{app:plots_dist}

\begin{figure}
    \centering
    \includegraphics[clip,trim=0 10 10 10,width=0.32\linewidth]{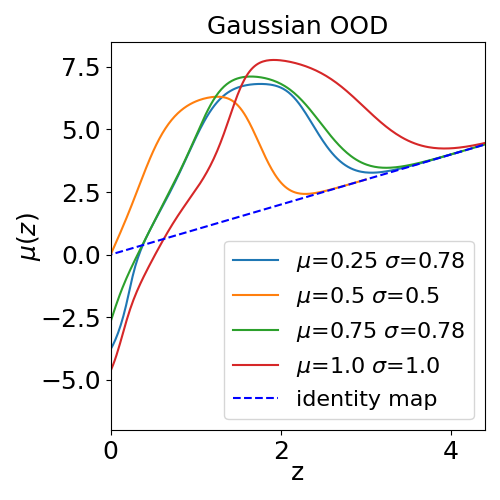}
    \includegraphics[clip,trim=0 10 10 10,width=0.32\linewidth]{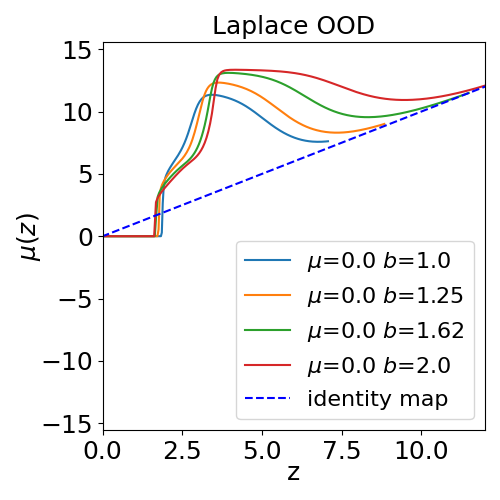}
    \includegraphics[clip,trim=0 10 10 10,width=0.32\linewidth]{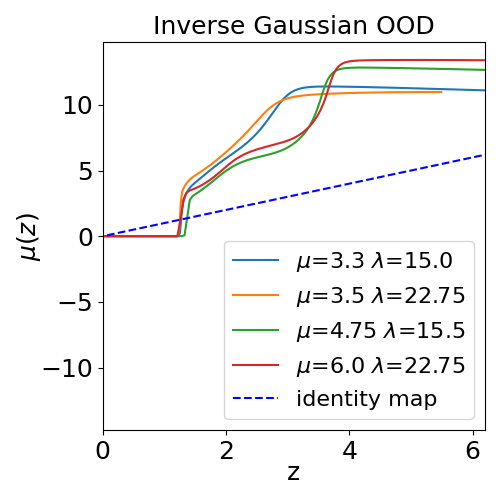}
    \caption{The mean of the OOD Gaussian Random Feature under the Gaussian (left), Laplace (middle) and Inverse Gaussian (right) distributions for the OOD distribution. Different curves on the same plot indicate different OOD distribution parameters. The weight on the IB term and weight of its $I(\tilde Z; Y)$ component are set as $\alpha=3.0$ and $\beta=10.0$. For the Gaussian case, $p(z|0)\sim \mathcal N(-0.5,0.5)$. For the Laplace case, $p(z|0)\sim \mathcal N(0,0.5)$. In the Inverse Gaussian case, $p(z|0)\sim \mathcal{N}(0,0.5)$. }
    \label{fig:OODfeature_diff_dist_params}
\end{figure}

In Figure~\ref{fig:OODfeature_diff_dist_params}, we explore how our feature shaping function varies with different distribution parameters.

\section{PLF Hyperparameter Tuning Using Bayesian Optimization}
\label{app: PLF_Bopt_settings}

As discussed in Section~\ref{sec: plf_motivation}, PLF involves a seven-parameter piecewise linear shaping function ~\citep{mondal2025variationalinformationtheoreticapproach}, where $f(\cdot)$ consists of three linear segments defined by breakpoints $z_1,z_2$, slopes $m_1,m_2$, and vertical offsets $y_{\text{start}}, y_{\text{end}}$ (with $y_1$ derived below). The breakpoints are parameterized by quantiles of the absolute ID feature distribution,
\[
z_1 = F^{-1}_{|Z_{\text{ID}}|}(q_1), \qquad
z_2 = F^{-1}_{|Z_{\text{ID}}|}(q_2), \qquad
q_2 = q_1 + \delta,
\]
where $F_{|Z_{\text{ID}}|}$ is the empirical cumulative distribution function of $|Z_{\text{ID}}|$. In practice, $z_1$ and $z_2$ are taken as the empirical order statistics of $|Z_{\text{ID}}|$ at ranks $\lfloor q_1 (n-1)\rfloor$ and $\lfloor q_2 (n-1)\rfloor$, where $n$ is the number of ID calibration samples.

The optimization variables are
\[
\theta = \left( y_{\text{start}}, y_{\text{end}}, \Delta y, q_1, u, m_1, m_2 \right),
\]
with bounds
\[
y_{\text{start}} \in [-2,2], \quad y_{\text{end}} \in [-2,2], \quad \Delta y \in [0,5],
\]
\[
q_1 \in [0.15,0.85], \quad u \in [0,1], \quad m_1 \in [0,2], \quad m_2 \in [-3,3].
\]
The value at the first breakpoint is $y_1 = y_{\text{end}} - \Delta y$, so that $\Delta y \geq 0$ corresponds to an upward jump from $y_1$ to $y_{\text{end}}$ at $x_1$.

Rather than optimizing $\delta$ directly, we optimize $u \in [0,1]$ and map it to $\delta$ via
\[
\delta = \delta_{\min} + u\,(\delta_{\max}-\delta_{\min}), \qquad
\delta_{\min}=0.10, \qquad \delta_{\max}=0.99-q_1,
\]
so that $q_2 = q_1 + \delta \in (q_1, 0.99]$ for any $u$. This keeps the search space a fixed axis-aligned box, since the admissible range of $\delta$ otherwise depends on $q_1$ through the constraint $q_2 \leq 0.99$.

Parameters are selected by Bayesian optimization with a Gaussian-process surrogate, implemented in \texttt{scikit-optimize}. We use $N$=100 total evaluations, of which the first $N_0$=80 are random initialization, with the \texttt{gp\_hedge} acquisition strategy. Each candidate $\theta$ is evaluated by applying the resulting shaping function to the validation ID and OOD features, passing the shaped features through the frozen classifier head to obtain energy scores, and computing the AUROC of ID-versus-OOD separation; the configuration maximizing this AUROC is retained.

\color{black}

\section{Empirical Insights Into Information Bottleneck Regularization}
\label{app:gaussian_noise_activations}

We study how IB regularization should be chosen with respect to properties of OOD data. This is important in practical scenarios. In particular, we conduct an experiment to suggest higher IB regularization is beneficial for ``noisier" OOD datasets.

\begin{figure}
    \centering
    \begin{subfigure}{0.19\textwidth}
        \centering
        \includegraphics[width=\textwidth]{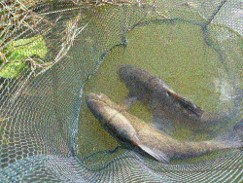}
        \caption{$\sigma = 25$}
    \end{subfigure}
    \begin{subfigure}{0.19\textwidth}
        \centering
        \includegraphics[width=\textwidth]{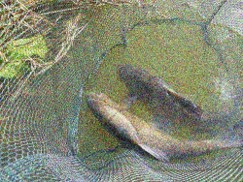}
        \caption{$\sigma = 50$}
    \end{subfigure}
    \begin{subfigure}{0.19\textwidth}
        \centering
        \includegraphics[width=\textwidth]{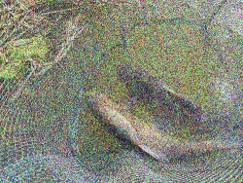}
        \caption{$\sigma = 100$}
    \end{subfigure}
    \begin{subfigure}{0.19\textwidth}
        \centering
        \includegraphics[width=\textwidth]{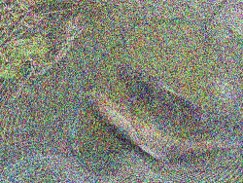}
        \caption{$\sigma = 150$}
    \end{subfigure}
    \begin{subfigure}{0.19\textwidth}
        \centering
        \includegraphics[width=\textwidth]{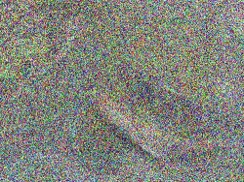}
        \caption{$\sigma = 255$}
    \end{subfigure}
    \caption{Visualization of a sample image from ImageNet validation split under different levels of noise corruption ($\sigma$ values).}
    \label{fig:sigma_comparison}
\end{figure}

\begin{figure}
    \centering
    \begin{subfigure}{0.49\textwidth}
    \centering
    \includegraphics[width=\textwidth]{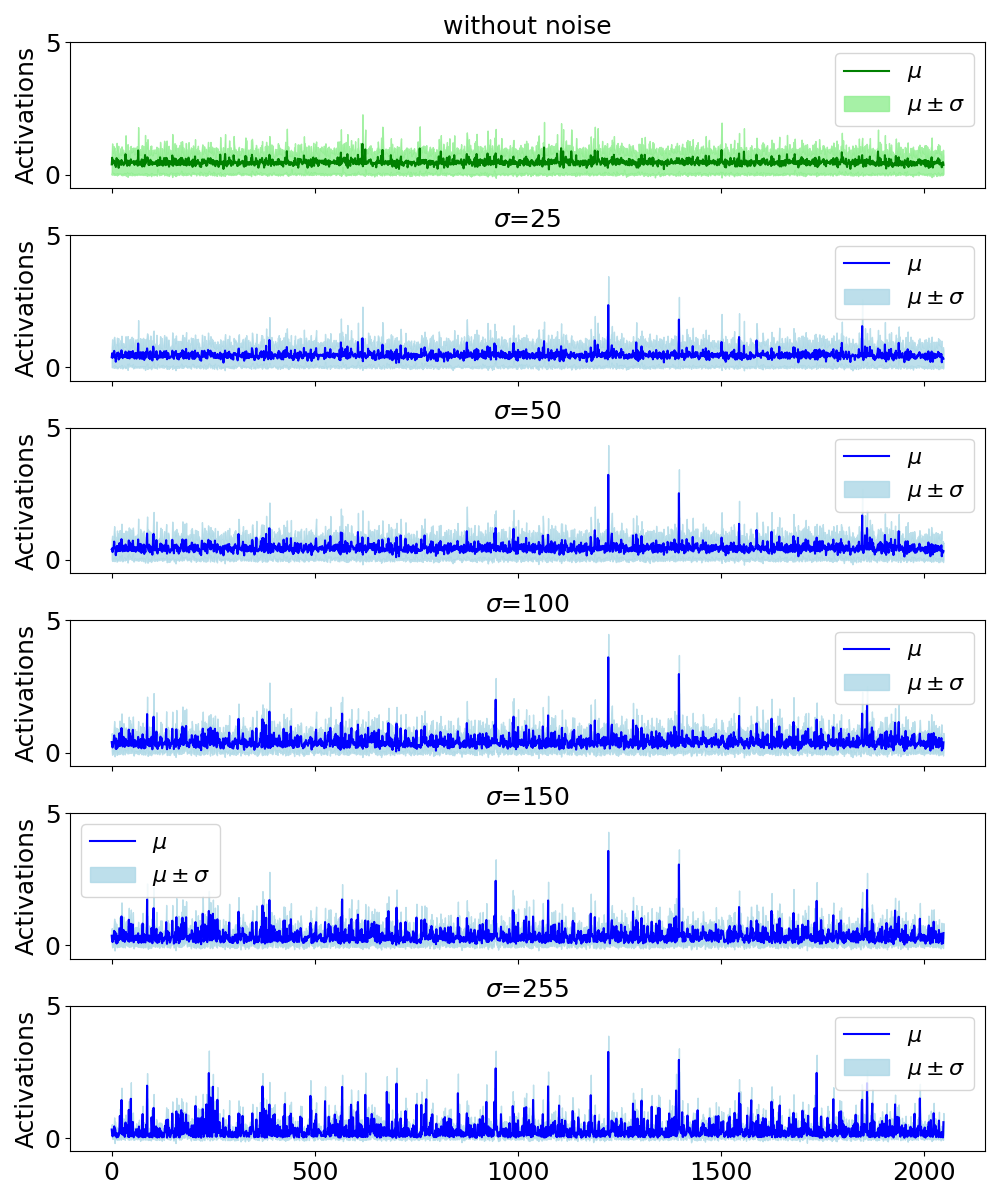}
    \caption{Activation range of $\mu \pm \sigma$}
    \label{fig:gaussian_noise_activations:a}
    \end{subfigure}
    \begin{subfigure}{0.49\textwidth}
    \centering
    \includegraphics[width=\linewidth]{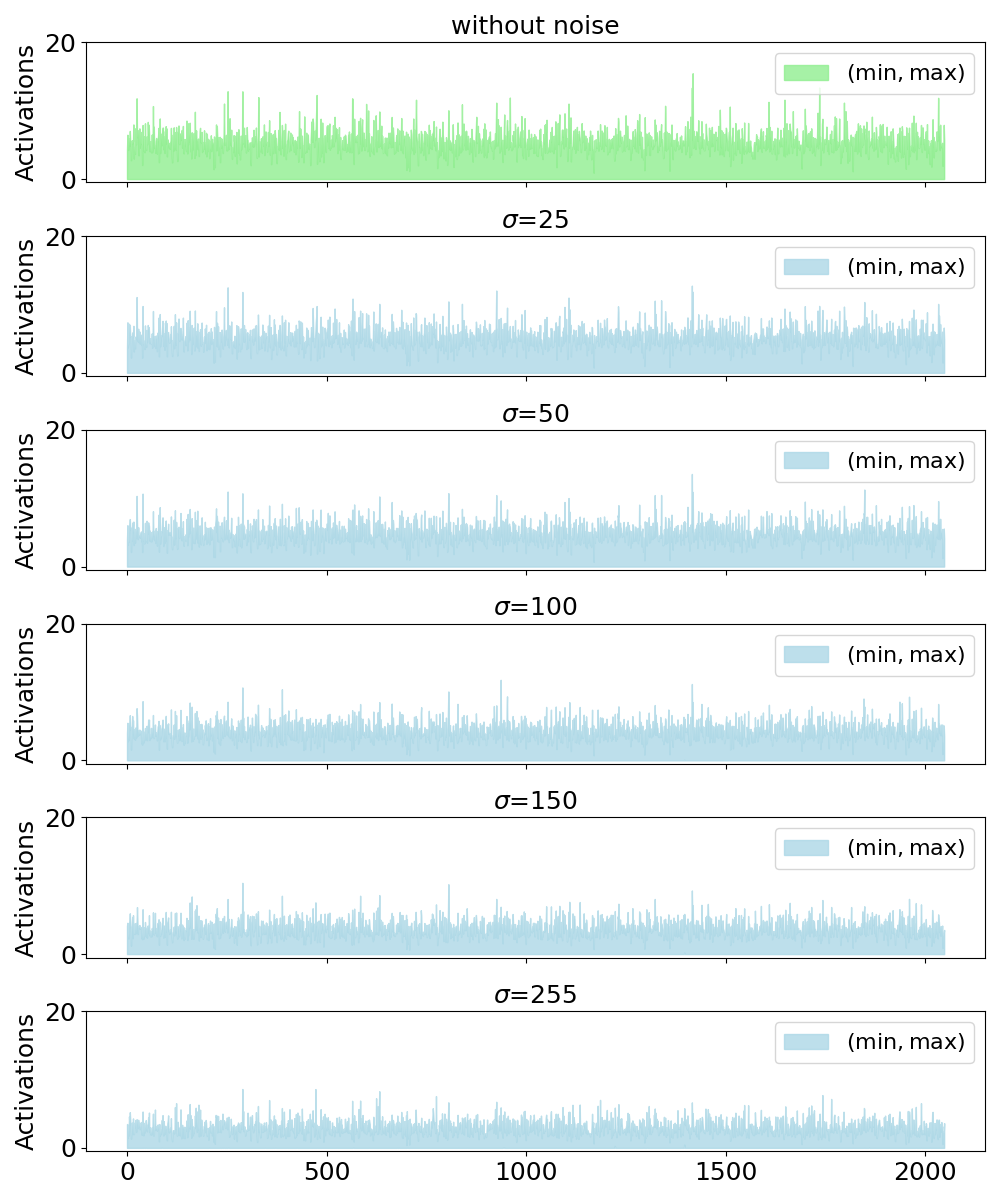}
    \caption{Activation range of ($\min$, $\max$)}
    \label{fig:gaussian_noise_activations:b}
    \end{subfigure}
    \caption{The network activations of ImageNet with different levels of additive Gaussian noises. The shaded regions in (a) represent one standard deviation above and below the mean, while those in (b) represent the range of min and max values of activations. With more noises, activations tend to more instable, smaller range of activation values.}
    \label{fig:gaussian_noise_activations}
\end{figure}

To this end, we conduct a series of controlled experiments using ResNet-50 trained on the ImageNet-1k dataset. We aim to determine the structure of optimal feature-shaping functions as a function of noise. We apply additive Gaussian noise $\mathcal{N}(0, \sigma)$ to the ImageNet-1k validation set and consider them as OOD data. The standard deviation $\sigma$ used are $\{25, 50, 100, 150, 255\}$ to create 5 OOD datasets. Visualizations of this data and activation patterns are shown in  Figure~\ref{fig:sigma_comparison} and Figure~\ref{fig:gaussian_noise_activations}.

Figure~\ref{fig:sigma_comparison} shows a visualization of an example ImageNet image and various noise levels added to it to simulate OOD data for the experiment. Figure~\ref{fig:gaussian_noise_activations} visualizes the activations of images from the validation set of ImageNet-1k with different levels of additive Gaussian noises. From figure~\ref{fig:gaussian_noise_activations:a}, with more noises, the activations are more instable and have more larger spikes. From figure~\ref{fig:gaussian_noise_activations:b}, the activations shrink more and peak more at small values. We observe that this data closely resembles the high variance of activation patterns in OOD datasets as noted in \citep{sun2021react}, and so our noisy data serves to mimic OOD data with varying noise levels. By examining how the learned features adapt under different noise levels, we gain insights into the relationship between the OOD data and the IB regularization term for optimal shaping.

\renewcommand{\thesubfigure}{\alph{subfigure}}

\begin{figure}
    \centering
    \includegraphics[width=0.8\linewidth]{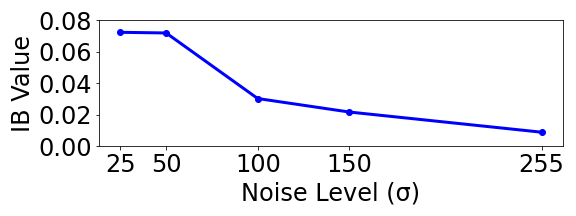}
    \caption{IB for the optimal hyperparameter optimized shaping function as a function of the noise level of OOD data. Lower IB corresponds to a more regularized shaping function.}
    \label{fig:IB_vs_noise}
\end{figure}

In Figure~\ref{fig:IB_vs_noise}, we plot the IB term of the obtained optimal shaping function optimized over hyperparameters at each noise level. Note we have used the Laplacian OOD distribution to estimate the IB term (Inverse Gaussian also results in similar results). It is seen that higher noise levels result in optimal shaping functions with lower IB values, which correspond to higher degree of regularization of the IB term in our loss functional. Thus, noisier OOD datasets require higher IB regularization for best OOD detection performance.

\section{Empirical Distribution of Network Feature Outputs}
\label{app:feature_distributions}
\begin{figure}
    \centering
    \includegraphics[width=0.9\linewidth]{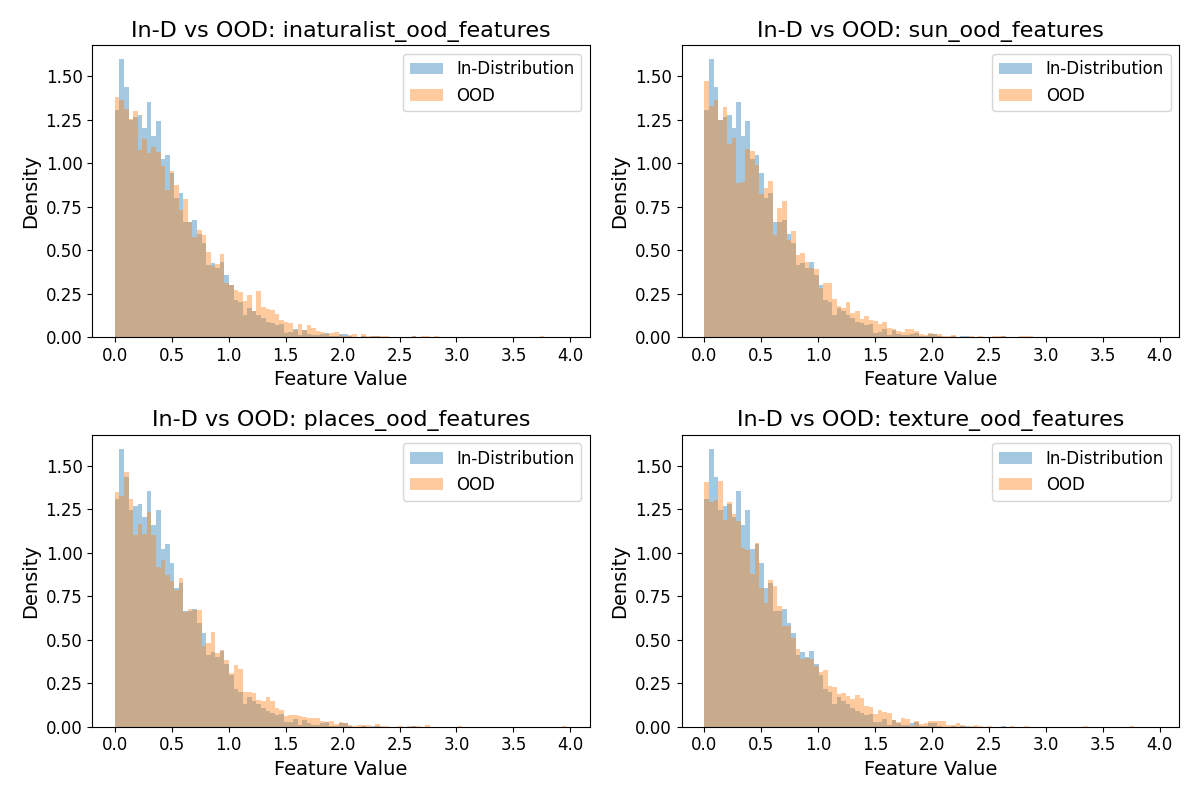}
    \caption{Distribution of features from the penultimate layer of ViT-B-16. Comparison with In-distribution data (ImageNet-1k) and different test OOD datasets in the ImageNet-1k benchmark~\citep{zhao2024towards}.  The ID and OOD  distributions resemble the positive part of a Gaussian with different variance.}
    \label{fig:vit_b_features}
\end{figure}

\begin{figure}
    \centering
    \includegraphics[width=0.9\linewidth]{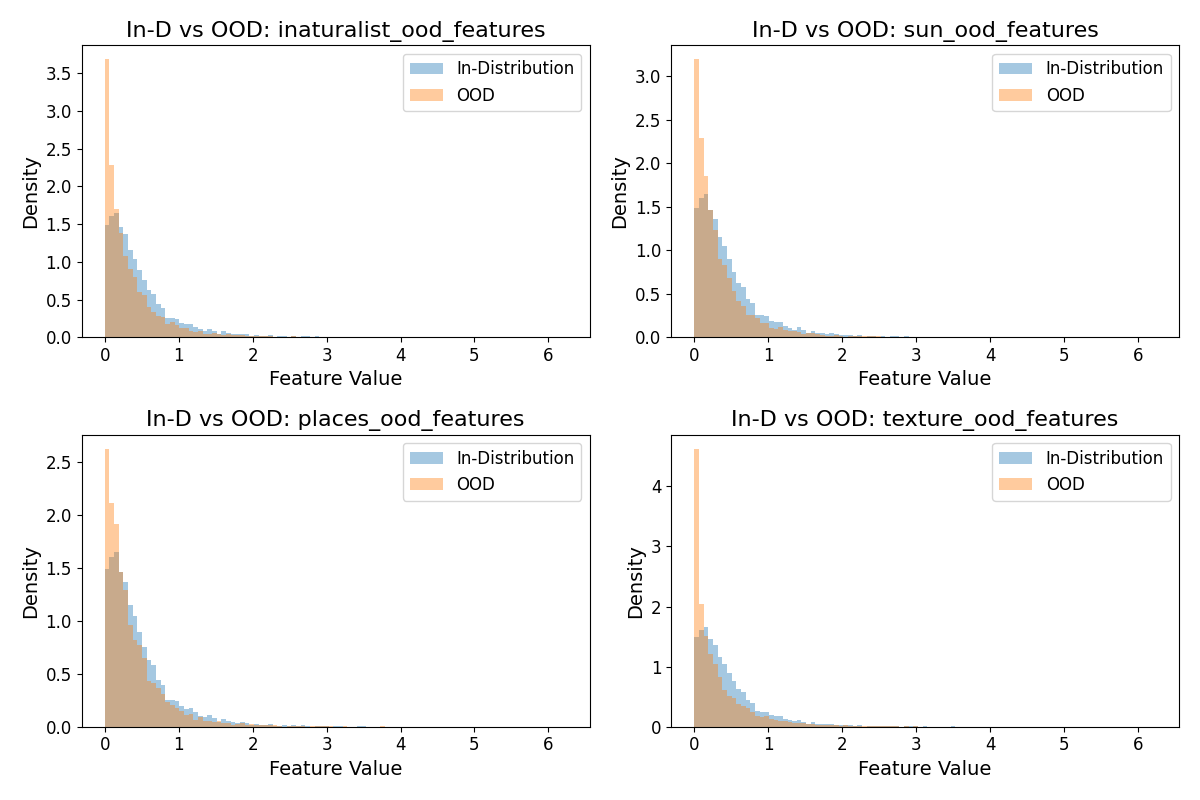}
    \caption{Distribution of features from the penultimate layer of ResNet-50. Comparison with In-distribution data (ImageNet-1k) and different test OOD datasets in the ImageNet-1k benchmark~\citep{zhao2024towards}.  The ID distribution resembles the positive part of a Gaussian and the OOD resembles the positive part of a Laplacian distribution.}
    \label{fig:resnet_features}
\end{figure}

\begin{figure}
    \centering
    \includegraphics[width=0.9\linewidth]{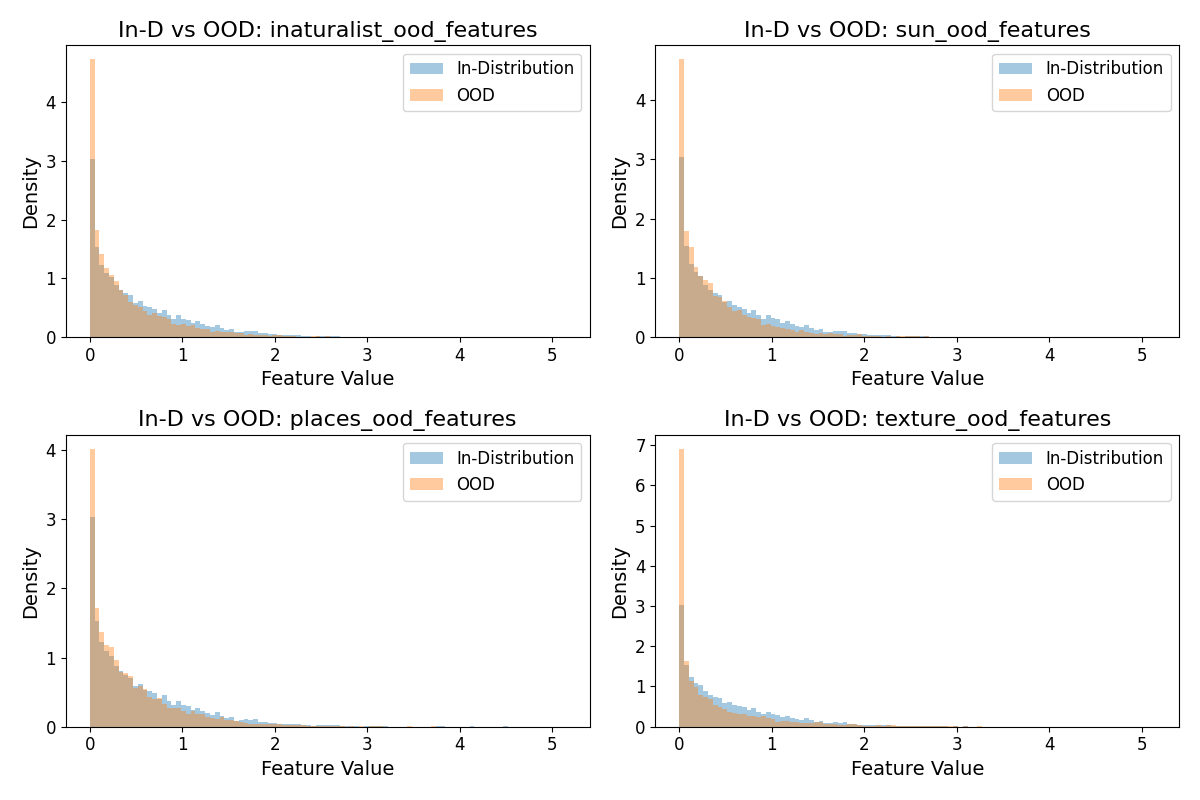}
    \caption{Distribution of features from the penultimate layer of MobileNet-V2. Comparison with In-distribution data (ImageNet-1k) and different test OOD datasets in the ImageNet-1k benchmark~\citep{zhao2024towards}. The ID and OOD both appear Laplacian; although this does not fit the ID assumption analyzed in this paper, our method nevertheless works well on this benchmark. }
    \label{fig:mobilenet_features}
\end{figure}

\begin{figure}
    \centering
    \includegraphics[width=0.9\linewidth]{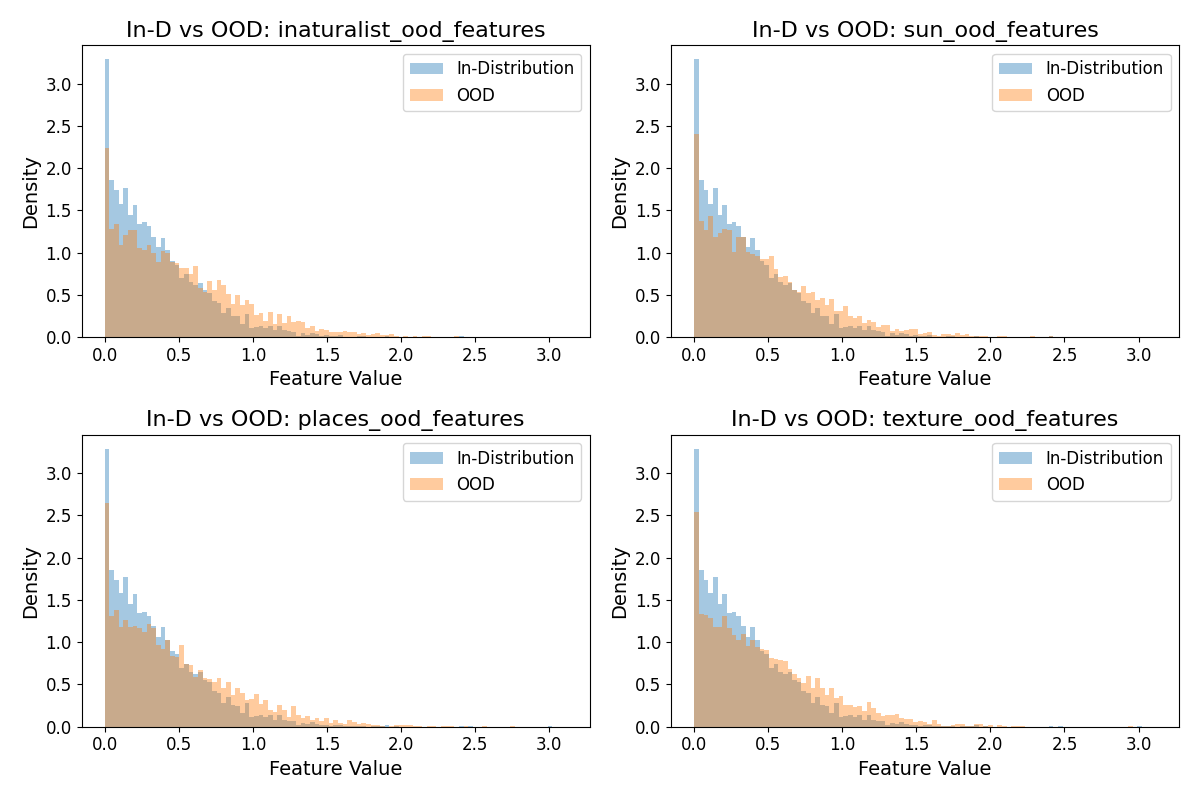}
    \caption{Distribution of features from the penultimate layer of ViT-L-16. Comparison with In-distribution data (ImageNet-1k) and different test OOD datasets in the ImageNet-1k benchmark~\citep{zhao2024towards}. The ID and OOD distributions appear Gaussian but with heavy weight on zeros; although these distributions don't fit the distributional assumptions in the cases analyzed in the paper, our method nevertheless performs well on this benchmark.}
    \label{fig:vit_L_features}
\end{figure}

In this section, we show the empirical distributions of ID and OOD data for various ImageNet-1k benchmarks.  As is shown in the subsequent figures, some benchmarks/architectures resemble the distribution assumptions analyzed in this paper (e.g., Gaussian ID/Gaussian OOD - Figure~\ref{fig:vit_b_features} and Gaussian ID/Laplacian OOD - Figure~\ref{fig:resnet_features} ), thus showing that these may be realistic assumptions. On the otherhand, some datasets (Figure~\ref{fig:mobilenet_features} and Figure~\ref{fig:vit_L_features}) exhibit distributions that do not fit the distributional assumptions analyzed in this paper.  Nevertheless, our novel shaping function still performs well on these benchmarks, showing that our shaping function may well work even when the data differs from the assumed distributions, which is important in practice as exact distributions may not be known.


\vskip 0.2in
\newpage
\bibliography{ood_lit}

@inproceedings{guille2024expecting,
  title     = {Expecting The Unexpected: Towards Broad Out-Of-Distribution Detection},
  author    = {Guille-Escuret, Charles and No{\"e}l, Pierre-Andr{\'e} and Mitliagkas, Ioannis and Vazquez, David and Monteiro, Joao},
  booktitle = {Advances in Neural Information Processing Systems (NeurIPS), Datasets and Benchmarks Track},
  volume    = {37},
  pages     = {130953--130976},
  year      = {2024},
  url       = {https://proceedings.neurips.cc/paper_files/paper/2024/file/ec80d18205e39d42a27192d5f3ddd688-Paper-Datasets_and_Benchmarks_Track.pdf}
}

@InProceedings{vaze2022openset,
               title={Open-Set Recognition: a Good Closed-Set Classifier is All You Need?},
               author={Sagar Vaze and Kai Han and Andrea Vedaldi and Andrew Zisserman},
               booktitle={International Conference on Learning Representations},
               year={2022}}

@article{krizhevsky2009learning,
  title={Learning multiple layers of features from tiny images},
  author={Krizhevsky, Alex and Hinton, Geoffrey and others},
  year={2009},
  publisher={Toronto, ON, Canada}
}

@inproceedings{imagenet_paper,
author = { Deng, Jia and Dong, Wei and Socher, Richard and Li, Li-Jia and Kai Li and Li Fei-Fei },
booktitle = { 2009 IEEE Computer Society Conference on Computer Vision and Pattern Recognition Workshops (CVPR Workshops) },
title = {{ ImageNet: A large-scale hierarchical image database }},
year = {2009},
volume = {},
ISSN = {1063-6919},
pages = {248-255},
doi = {10.1109/CVPR.2009.5206848},
url = {https://doi.ieeecomputersociety.org/10.1109/CVPR.2009.5206848},
publisher = {IEEE Computer Society},
address = {Los Alamitos, CA, USA},
month =Jun}

@inproceedings{imagenet-v2,
  title={Do ImageNet Classifiers Generalize to ImageNet?},
  author={Recht, Benjamin and Roelofs, Rebecca and Schmidt, Ludwig and Shankar, Vaishaal},
  booktitle={Proceedings of the 36th International Conference on Machine Learning},
  volume={97},
  year={2019},
  organization={PMLR},
  url={https://proceedings.mlr.press/v97/recht19a.html}
}

@inproceedings{imagenet-r,
  title={The Many Faces of Robustness: A Critical Analysis of Out-of-Distribution Generalization},
  author={Hendrycks, Dan and Basart, Steven and Mu, Norman and Kadavath, Saurav and Wang, Frank and Dorundo, Evan and Desai, Rahul and Zhu, Tyler and Parajuli, Samyak and Guo, Mike and Song, Dawn and Steinhardt, Jacob and Gilmer, Justin},
  booktitle={Proceedings of the IEEE/CVF International Conference on Computer Vision (ICCV)},
  year={2021},
  url={https://openaccess.thecvf.com/content/ICCV2021/html/Hendrycks_The_Many_Faces_of_Robustness_A_Critical_Analysis_of_Out-of-Distribution_ICCV_2021_paper.html}
}

@inproceedings{imagenet-c,
  title={Benchmarking Neural Network Robustness to Common Corruptions and Perturbations},
  author={Hendrycks, Dan and Dietterich, Thomas},
  booktitle={International Conference on Learning Representations},
  year={2019},
  url={https://openreview.net/forum?id=HJz6tiCqYm}
}

@inproceedings{he2015deepresiduallearningimage,
  title={Deep Residual Learning for Image Recognition},
  author={He, Kaiming and Zhang, Xiangyu and Ren, Shaoqing and Sun, Jian},
  booktitle={Proceedings of the IEEE Conference on Computer Vision and Pattern Recognition (CVPR)},
  pages={770--778},
  year={2016},
  doi={10.1109/CVPR.2016.90},
  url={https://openaccess.thecvf.com/content_cvpr_2016/html/He_Deep_Residual_Learning_CVPR_2016_paper.html}
}

@misc{krizhevsky2009cifar,
  title        = {CIFAR-10 and CIFAR-100 Datasets},
  author       = {Krizhevsky, Alex and Nair, Vinod and Hinton, Geoffrey},
  year         = {2009},
  howpublished = {\url{https://www.cs.toronto.edu/~kriz/cifar.html}},
  note         = {Accessed: 2026-05-02}
}

@misc{madry2019deeplearningmodelsresistant,
      title={Towards Deep Learning Models Resistant to Adversarial Attacks}, 
      author={Aleksander Madry and Aleksandar Makelov and Ludwig Schmidt and Dimitris Tsipras and Adrian Vladu},
      year={2019},
      eprint={1706.06083},
      archivePrefix={arXiv},
      primaryClass={stat.ML},
      url={https://arxiv.org/abs/1706.06083}, 
}

@InProceedings{autoattack,
  title = 	 {Reliable evaluation of adversarial robustness with an ensemble of diverse parameter-free attacks},
  author =       {Croce, Francesco and Hein, Matthias},
  booktitle = 	 {Proceedings of the 37th International Conference on Machine Learning},
  pages = 	 {2206--2216},
  year = 	 {2020},
  editor = 	 {III, Hal Daumé and Singh, Aarti},
  volume = 	 {119},
  series = 	 {Proceedings of Machine Learning Research},
  month = 	 {13--18 Jul},
  publisher =    {PMLR},
  url = 	 {https://proceedings.mlr.press/v119/croce20b.html},

}

@article{cBigGAN,
  author       = {Andrew Brock and
                  Jeff Donahue and
                  Karen Simonyan},
  title        = {Large Scale {GAN} Training for High Fidelity Natural Image Synthesis},
  journal      = {CoRR},
  volume       = {abs/1809.11096},
  year         = {2018},
  url          = {http://arxiv.org/abs/1809.11096},
  eprinttype   = {arXiv},
  eprint       = {1809.11096},
  bibsource    = {dblp computer science bibliography, https://dblp.org}
}

@INPROCEEDINGS {diffusion,
author = { Rombach, Robin and Blattmann, Andreas and Lorenz, Dominik and Esser, Patrick and Ommer, Bjorn },
booktitle = { 2022 IEEE/CVF Conference on Computer Vision and Pattern Recognition (CVPR) },
title = {{ High-Resolution Image Synthesis with Latent Diffusion Models }},
year = {2022},
volume = {},
ISSN = {},
pages = {10674-10685},
doi = {10.1109/CVPR52688.2022.01042},
url = {https://doi.ieeecomputersociety.org/10.1109/CVPR52688.2022.01042},
publisher = {IEEE Computer Society},
address = {Los Alamitos, CA, USA},
month =Jun}

@misc{bitterwolf2023outfixingimagenetoutofdistribution,
      title={In or Out? Fixing ImageNet Out-of-Distribution Detection Evaluation}, 
      author={Julian Bitterwolf and Maximilian Müller and Matthias Hein},
      year={2023},
      eprint={2306.00826},
      archivePrefix={arXiv},
      primaryClass={cs.LG},
      url={https://arxiv.org/abs/2306.00826}, 
}

@article{sun2021react,
  title={React: Out-of-distribution detection with rectified activations},
  author={Sun, Yiyou and Guo, Chuan and Li, Yixuan},
  journal={Advances in Neural Information Processing Systems},
  volume={34},
  pages={144--157},
  year={2021}
}

@article{zhao2024towards,
  title={Towards Optimal Feature-Shaping Methods for Out-of-Distribution Detection},
  author={Zhao, Qinyu and Xu, Ming and Gupta, Kartik and Asthana, Akshay and Zheng, Liang and Gould, Stephen},
  journal={arXiv preprint arXiv:2402.00865},
  year={2024}
}

@misc{xu2023vravariationalrectifiedactivation,
      title={VRA: Variational Rectified Activation for Out-of-distribution Detection}, 
      author={Mingyu Xu and Zheng Lian and Bin Liu and Jianhua Tao},
      year={2023},
      eprint={2302.11716},
      archivePrefix={arXiv},
      primaryClass={cs.LG},
      url={https://arxiv.org/abs/2302.11716}, 
}

@book{cover1999elements,
  title={Elements of information theory},
  author={Cover, Thomas M},
  year={1999},
  publisher={John Wiley \& Sons}
}

@article{tishby2000information,
  title={The information bottleneck method},
  author={Tishby, Naftali and Pereira, Fernando C and Bialek, William},
  journal={arXiv preprint physics/0004057},
  year={2000}
}

@misc{frazier2018tutorialbayesianoptimization,
      title={A Tutorial on Bayesian Optimization}, 
      author={Peter I. Frazier},
      year={2018},
      eprint={1807.02811},
      archivePrefix={arXiv},
      primaryClass={stat.ML},
      url={https://arxiv.org/abs/1807.02811}, 
}

@book{troutman2012variational,
  title={Variational calculus and optimal control: optimization with elementary convexity},
  author={Troutman, John L},
  year={2012},
  publisher={Springer Science \& Business Media}
}

@misc{msp_hendrycks,
      title={A Baseline for Detecting Misclassified and Out-of-Distribution Examples in Neural Networks}, 
      author={Dan Hendrycks and Kevin Gimpel},
      year={2018},
      eprint={1610.02136},
      archivePrefix={arXiv},
      primaryClass={cs.NE},
      url={https://arxiv.org/abs/1610.02136}, 
}

@misc{odin_liang,
      title={Enhancing The Reliability of Out-of-distribution Image Detection in Neural Networks}, 
      author={Shiyu Liang and Yixuan Li and R. Srikant},
      year={2020},
      eprint={1706.02690},
      archivePrefix={arXiv},
      primaryClass={cs.LG},
      url={https://arxiv.org/abs/1706.02690}, 
}

@InProceedings{Resnet50,
author="He, Kaiming
and Zhang, Xiangyu
and Ren, Shaoqing
and Sun, Jian",
editor="Leibe, Bastian
and Matas, Jiri
and Sebe, Nicu
and Welling, Max",
title="Identity Mappings in Deep Residual Networks",
booktitle="Computer Vision -- ECCV 2016",
year="2016",
publisher="Springer International Publishing",
address="Cham",
pages="630--645",
}

@InProceedings{Kong_BFAct,
author="Kong, Haojia
and Li, Haoan",
editor="Sun, Fuchun
and Cangelosi, Angelo
and Zhang, Jianwei
and Yu, Yuanlong
and Liu, Huaping
and Fang, Bin",
title="BFAct: Out-of-Distribution Detection with Butterworth Filter Rectified Activations",
booktitle="Cognitive Systems and Information Processing",
year="2023",
publisher="Springer Nature Singapore",
address="Singapore",
pages="115--129"
}

@article{djurisic2022ash,
    url = {https://arxiv.org/abs/2209.09858},
    author = {Djurisic, Andrija and Bozanic, Nebojsa and Ashok, Arjun and Liu, Rosanne},
    title = {Extremely Simple Activation Shaping for Out-of-Distribution Detection},
    publisher = {arXiv},
    year = {2022},
    }

@misc{devries2018learning,
      title={Learning Confidence for Out-of-Distribution Detection in Neural Networks}, 
      author={Terrance DeVries and Graham W. Taylor},
      year={2018},
      eprint={1802.04865},
      archivePrefix={arXiv},
      primaryClass={stat.ML}
}

@misc{yang2022generalized,
      title={Generalized Out-of-Distribution Detection: A Survey}, 
      author={Jingkang Yang and Kaiyang Zhou and Yixuan Li and Ziwei Liu},
      year={2022},
      eprint={2110.11334},
      archivePrefix={arXiv},
      primaryClass={cs.CV}
}

@misc{ahn2023line,
      title={LINe: Out-of-Distribution Detection by Leveraging Important Neurons}, 
      author={Yong Hyun Ahn and Gyeong-Moon Park and Seong Tae Kim},
      year={2023},
      eprint={2303.13995},
      archivePrefix={arXiv},
      primaryClass={cs.CV}
}

@inproceedings{sun2022dice,
  title={DICE: Leveraging Sparsification for Out-of-Distribution Detection},
  author={Sun, Yiyou and Li, Yixuan},
  booktitle={European Conference on Computer Vision},
  year={2022}
}

@inproceedings{zhanglearning,
  title={Learning to Shape In-distribution Feature Space for Out-of-distribution Detection},
  author={Zhang, Yonggang and Lu, Jie and Peng, Bo and Fang, Zhen and Cheung, Yiu-ming},
  booktitle={The Thirty-eighth Annual Conference on Neural Information Processing Systems}
}

@inproceedings{
zhang2024learning,
title={Learning to Shape In-distribution Feature Space for Out-of-distribution Detection},
author={Yonggang Zhang and Jie Lu and Bo Peng and Zhen Fang and Yiu-ming Cheung},
booktitle={The Thirty-eighth Annual Conference on Neural Information Processing Systems},
year={2024},
url={https://openreview.net/forum?id=1Du3mMP5YN}
}

@misc{hendrycks2018,
      title={A Baseline for Detecting Misclassified and Out-of-Distribution Examples in Neural Networks}, 
      author={Dan Hendrycks and Kevin Gimpel},
      year={2018},
      eprint={1610.02136},
      archivePrefix={arXiv},
      primaryClass={cs.NE},
      url={https://arxiv.org/abs/1610.02136}, 
}

@INPROCEEDINGS{zhang_decoupling,
  author={Zhang, Zihan and Xiang, Xiang},
  booktitle={2023 IEEE/CVF Conference on Computer Vision and Pattern Recognition (CVPR)}, 
  title={Decoupling MaxLogit for Out-of-Distribution Detection}, 
  year={2023},
  volume={},
  number={},
  pages={3388-3397},
  doi={10.1109/CVPR52729.2023.00330}}

@misc{liu2021energy,
      title={Energy-based Out-of-distribution Detection}, 
      author={Weitang Liu and Xiaoyun Wang and John D. Owens and Yixuan Li},
      year={2021},
      eprint={2010.03759},
      archivePrefix={arXiv},
      primaryClass={cs.LG},
      url={https://arxiv.org/abs/2010.03759}, 
}

@article{wu2023energy,
  title={Energy-based out-of-distribution detection for graph neural networks},
  author={Wu, Qitian and Chen, Yiting and Yang, Chenxiao and Yan, Junchi},
  journal={arXiv preprint arXiv:2302.02914},
  year={2023}
}

@article{elflein2021out,
  title={On out-of-distribution detection with energy-based models},
  author={Elflein, Sven and Charpentier, Bertrand and Z{\"u}gner, Daniel and G{\"u}nnemann, Stephan},
  journal={arXiv preprint arXiv:2107.08785},
  year={2021}
}

@article{lee2018simple,
  title={A simple unified framework for detecting out-of-distribution samples and adversarial attacks},
  author={Lee, Kimin and Lee, Kibok and Lee, Honglak and Shin, Jinwoo},
  journal={Advances in neural information processing systems},
  volume={31},
  year={2018}
}

@inproceedings{sun2022knn,
  title={Out-of-distribution detection with deep nearest neighbors},
  author={Sun, Yiyou and Ming, Yifei and Zhu, Xiaojin and Li, Yixuan},
  booktitle={International Conference on Machine Learning},
  pages={20827--20840},
  year={2022},
  organization={PMLR}
}

@misc{hendrycks2022_species,
      title={Scaling Out-of-Distribution Detection for Real-World Settings}, 
      author={Dan Hendrycks and Steven Basart and Mantas Mazeika and Andy Zou and Joe Kwon and Mohammadreza Mostajabi and Jacob Steinhardt and Dawn Song},
      year={2022},
      eprint={1911.11132},
      archivePrefix={arXiv},
      primaryClass={cs.CV},
      url={https://arxiv.org/abs/1911.11132}, 
}

@misc{vanhorn2018inaturalistspeciesclassificationdetection,
      title={The iNaturalist Species Classification and Detection Dataset}, 
      author={Grant Van Horn and Oisin Mac Aodha and Yang Song and Yin Cui and Chen Sun and Alex Shepard and Hartwig Adam and Pietro Perona and Serge Belongie},
      year={2018},
      eprint={1707.06642},
      archivePrefix={arXiv},
      primaryClass={cs.CV},
      url={https://arxiv.org/abs/1707.06642}, 
}

@INPROCEEDINGS{SUN_data,
  author={Xiao, Jianxiong and Hays, James and Ehinger, Krista A. and Oliva, Aude and Torralba, Antonio},
  booktitle={2010 IEEE Computer Society Conference on Computer Vision and Pattern Recognition}, 
  title={SUN database: Large-scale scene recognition from abbey to zoo}, 
  year={2010},
  volume={},
  number={},
  pages={3485-3492},
  doi={10.1109/CVPR.2010.5539970}}

@ARTICLE{places,
  author={Zhou, Bolei and Lapedriza, Agata and Khosla, Aditya and Oliva, Aude and Torralba, Antonio},
  journal={IEEE Transactions on Pattern Analysis and Machine Intelligence}, 
  title={Places: A 10 Million Image Database for Scene Recognition}, 
  year={2018},
  volume={40},
  number={6},
  pages={1452-1464},
  doi={10.1109/TPAMI.2017.2723009}}

@misc{wang2022_OpenImageO,
      title={ViM: Out-Of-Distribution with Virtual-logit Matching}, 
      author={Haoqi Wang and Zhizhong Li and Litong Feng and Wayne Zhang},
      year={2022},
      eprint={2203.10807},
      archivePrefix={arXiv},
      primaryClass={cs.CV},
      url={https://arxiv.org/abs/2203.10807}, 
}

@InProceedings{Hendrycks_2021_Imagenet-O,
    author    = {Hendrycks, Dan and Zhao, Kevin and Basart, Steven and Steinhardt, Jacob and Song, Dawn},
    title     = {Natural Adversarial Examples},
    booktitle = {Proceedings of the IEEE/CVF Conference on Computer Vision and Pattern Recognition (CVPR)},
    month     = {June},
    year      = {2021},
    pages     = {15262-15271}
}

@INPROCEEDINGS{cimpoi_textures,
  author={Cimpoi, Mircea and Maji, Subhransu and Kokkinos, Iasonas and Mohamed, Sammy and Vedaldi, Andrea},
  booktitle={2014 IEEE Conference on Computer Vision and Pattern Recognition}, 
  title={Describing Textures in the Wild}, 
  year={2014},
  volume={},
  number={},
  pages={3606-3613},
  doi={10.1109/CVPR.2014.461}}

@ARTICLE{mnist_deng,
  author={Deng, Li},
  journal={IEEE Signal Processing Magazine}, 
  title={The MNIST Database of Handwritten Digit Images for Machine Learning Research [Best of the Web]}, 
  year={2012},
  volume={29},
  number={6},
  pages={141-142},
  doi={10.1109/MSP.2012.2211477}}

@ARTICLE{tiny_imagenet,
  author={Torralba, Antonio and Fergus, Rob and Freeman, William T.},
  journal={IEEE Transactions on Pattern Analysis and Machine Intelligence}, 
  title={80 Million Tiny Images: A Large Data Set for Nonparametric Object and Scene Recognition}, 
  year={2008},
  volume={30},
  number={11},
  pages={1958-1970},
  doi={10.1109/TPAMI.2008.128}}

@inproceedings{svhn,
title	= {Reading Digits in Natural Images with Unsupervised Feature Learning},author	= {Yuval Netzer and Tao Wang and Adam Coates and Alessandro Bissacco and Bo Wu and Andrew Y. Ng},year	= {2011},URL	= {http://ufldl.stanford.edu/housenumbers/nips2011_housenumbers.pdf},booktitle	= {NIPS Workshop on Deep Learning and Unsupervised Feature Learning 2011}}

@misc{yu2016_lsun_cropped,
      title={LSUN: Construction of a Large-scale Image Dataset using Deep Learning with Humans in the Loop}, 
      author={Fisher Yu and Ari Seff and Yinda Zhang and Shuran Song and Thomas Funkhouser and Jianxiong Xiao},
      year={2016},
      eprint={1506.03365},
      archivePrefix={arXiv},
      primaryClass={cs.CV},
      url={https://arxiv.org/abs/1506.03365}, 
}

@misc{xu2015_isun,
      title={TurkerGaze: Crowdsourcing Saliency with Webcam based Eye Tracking}, 
      author={Pingmei Xu and Krista A Ehinger and Yinda Zhang and Adam Finkelstein and Sanjeev R. Kulkarni and Jianxiong Xiao},
      year={2015},
      eprint={1504.06755},
      archivePrefix={arXiv},
      primaryClass={cs.CV},
      url={https://arxiv.org/abs/1504.06755}, 
}

@inproceedings{Fort,
 author = {Fort, Stanislav and Ren, Jie and Lakshminarayanan, Balaji},
 booktitle = {Advances in Neural Information Processing Systems},
 editor = {M. Ranzato and A. Beygelzimer and Y. Dauphin and P.S. Liang and J. Wortman Vaughan},
 pages = {7068--7081},
 publisher = {Curran Associates, Inc.},
 title = {Exploring the Limits of Out-of-Distribution Detection},
 url = {https://proceedings.neurips.cc/paper_files/paper/2021/file/3941c4358616274ac2436eacf67fae05-Paper.pdf},
 volume = {34},
 year = {2021}
}

@INPROCEEDINGS{mobilenetv2,
  author={Sandler, Mark and Howard, Andrew and Zhu, Menglong and Zhmoginov, Andrey and Chen, Liang-Chieh},
  booktitle={2018 IEEE/CVF Conference on Computer Vision and Pattern Recognition}, 
  title={MobileNetV2: Inverted Residuals and Linear Bottlenecks}, 
  year={2018},
  volume={},
  number={},
  pages={4510-4520},
  doi={10.1109/CVPR.2018.00474}}

@misc{dosovitskiy2021_VIT_B,
      title={An Image is Worth 16x16 Words: Transformers for Image Recognition at Scale}, 
      author={Alexey Dosovitskiy and Lucas Beyer and Alexander Kolesnikov and Dirk Weissenborn and Xiaohua Zhai and Thomas Unterthiner and Mostafa Dehghani and Matthias Minderer and Georg Heigold and Sylvain Gelly and Jakob Uszkoreit and Neil Houlsby},
      year={2021},
      eprint={2010.11929},
      archivePrefix={arXiv},
      primaryClass={cs.CV},
      url={https://arxiv.org/abs/2010.11929}, 
}

@misc{russakovsky2015imagenet,
      title={ImageNet Large Scale Visual Recognition Challenge}, 
      author={Olga Russakovsky and Jia Deng and Hao Su and Jonathan Krause and Sanjeev Satheesh and Sean Ma and Zhiheng Huang and Andrej Karpathy and Aditya Khosla and Michael Bernstein and Alexander C. Berg and Li Fei-Fei},
      year={2015},
      eprint={1409.0575},
      archivePrefix={arXiv},
      primaryClass={cs.CV},
      url={https://arxiv.org/abs/1409.0575}, 
}

@article{ren2019likelihood,
  title={Likelihood ratios for out-of-distribution detection},
  author={Ren, Jie and Liu, Peter J and Fertig, Emily and Snoek, Jasper and Poplin, Ryan and Depristo, Mark and Dillon, Joshua and Lakshminarayanan, Balaji},
  journal={Advances in neural information processing systems},
  volume={32},
  year={2019}
}

@article{kirichenko2020normalizing,
  title={Why normalizing flows fail to detect out-of-distribution data},
  author={Kirichenko, Polina and Izmailov, Pavel and Wilson, Andrew G},
  journal={Advances in neural information processing systems},
  volume={33},
  pages={20578--20589},
  year={2020}
}

@article{fort2021exploring,
  title={Exploring the limits of out-of-distribution detection},
  author={Fort, Stanislav and Ren, Jie and Lakshminarayanan, Balaji},
  journal={Advances in neural information processing systems},
  volume={34},
  pages={7068--7081},
  year={2021}
}

@misc{zhang2024openoodv15enhancedbenchmark,
      title={OpenOOD v1.5: Enhanced Benchmark for Out-of-Distribution Detection}, 
      author={Jingyang Zhang and Jingkang Yang and Pengyun Wang and Haoqi Wang and Yueqian Lin and Haoran Zhang and Yiyou Sun and Xuefeng Du and Yixuan Li and Ziwei Liu and Yiran Chen and Hai Li},
      year={2024},
      eprint={2306.09301},
      archivePrefix={arXiv},
      primaryClass={cs.LG},
      url={https://arxiv.org/abs/2306.09301}, 
}

@misc{mondal2025variationalinformationtheoreticapproach,
      title={A Variational Information Theoretic Approach to Out-of-Distribution Detection}, 
      author={Sudeepta Mondal and Zhuolin Jiang and Ganesh Sundaramoorthi},
      year={2025},
      eprint={2506.14194},
      archivePrefix={arXiv},
      primaryClass={cs.LG},
      url={https://arxiv.org/abs/2506.14194}, 
}

@article{magesh2023principled,
  title={Principled out-of-distribution detection via multiple testing},
  author={Magesh, Akshayaa and Veeravalli, Venugopal V and Roy, Anirban and Jha, Susmit},
  journal={Journal of Machine Learning Research},
  volume={24},
  number={378},
  pages={1--35},
  year={2023}
}

@article{yang2023full,
  title={Full-spectrum out-of-distribution detection},
  author={Yang, Jingkang and Zhou, Kaiyang and Liu, Ziwei},
  journal={International Journal of Computer Vision},
  volume={131},
  number={10},
  pages={2607--2622},
  year={2023},
  publisher={Springer}
}

@inproceedings{yang2024imagenet,
  title={Imagenet-ood: Deciphering modern out-of-distribution detection algorithms},
  author={Yang, William and Zhang, Byron and Russakovsky, Olga},
  booktitle={International Conference on Learning Representations},
  volume={2024},
  pages={41877--41906},
  year={2024}
}

\end{document}